%% file: main.tex
\documentclass{IEEEtran}
\input{parts/preamble}

\begin{document}
\input{parts/frontmatter}
\input{parts/introduction}

\input{parts/data}

\input{parts/machine_learning}
\input{parts/results}

\input{parts/conclusions}

\input{parts/acknowledgments}
\printglossaries
\bibliographystyle{IEEEtran}
\bibliography{references.bib}
\input{parts/supplementary}
\end{document}

%% file: parts/preamble.tex
\usepackage{graphicx}
\graphicspath{ {./img/} }
\usepackage{amssymb}
\usepackage[font=small]{caption}
\usepackage{subcaption}
\usepackage[inline]{enumitem}
\usepackage{glossaries}
\usepackage{color}
\usepackage{xcolor}
\usepackage[export]{adjustbox}
\PassOptionsToPackage{hyphens}{url}\usepackage{hyperref}
\loadglsentries{parts/glossories.tex}
\usepackage[noadjust]{cite}

\newcommand{\npos}{318}
\newcommand{\nneg}{1686}
\newcommand{\ntot}{2004}
\newcommand{\ntest}{435}

%% file: parts/glossories.tex
\newacronym{sar}{SAR}{synthetic aperture radar}
\newacronym{cnn}{CNN}{convolutional neural network}
\newacronym{ecmwf}{ECMWF}{european centre for medium-range weather forecasts}
\newacronym{era5}{ERA5}{ECMWF reanalysis version 5}
\newacronym{erai}{ERA-I}{ECMWF reanalysis Interim}
\newacronym{slp}{SLP}{sea level pressure}
\newacronym{aoi}{AOI}{area of interest}
\newacronym{asf}{ASF}{Alaska satellite facility}
\newacronym{utm}{UTM}{universal transverse mercator}
\newacronym{rgb}{RGB}{red-green-blue}
\newacronym{ew}{EW}{extended wide-swath}
\newacronym{iw}{IW}{interferometric wide-swath}
\newacronym{grd}{GRD}{ground range detected}
\newacronym{sep2d}{Sep2DConv}{separable 2D convolutions}
\newacronym{ig}{IG}{integrated gradients}
\newacronym{gcam}{Grad-CAM}{gradient-weighted class activation mapping}
\newacronym{nwp}{NWP}{numerical weather prediction}
\newacronym{ir}{IR}{infra-red}
\newacronym{tn}{TN}{true negative}
\newacronym{tp}{TP}{true positive}
\newacronym{fn}{FN}{false negative}
\newacronym{fp}{FP}{false positive}
\newacronym{esa}{ESA}{European space agency}
\newacronym{gdar}{GDAR}{Generic DAta Raster}
\newacronym{norce}{NORCE}{Norwegian Research Center}
\newacronym{stars}{STARS}{Sea Surface Temperature and Altimeter Synergy for Improved Forecasting of Polar Lows}

%% file: parts/frontmatter.tex



\title{Recognition of polar lows in Sentinel-1 SAR images with deep learning
}

\author{Jakob Grahn$^{*}$,
        Filippo Maria Bianchi
\thanks{*jgra@norceresearch.no}
\thanks{J. Grahn is with NORCE, The Norwegian Research Centre AS}
\thanks{F. M. Bianchi is with the Dept. of Mathematics and Statistics, UiT the Arctic University of Norway and with NORCE, The Norwegian Research Centre AS}%
}

\maketitle

\begin{abstract}
In this paper, we explore the possibility of detecting polar lows in C-band \acrshort{sar} images by means of deep learning. Specifically, we introduce a novel dataset consisting of Sentinel-1 images divided into two classes, representing the \emph{presence} and \emph{absence} of a maritime mesocyclone, respectively. 
The dataset is constructed using the \acrshort{era5} dataset as baseline and it consists of \ntot{} annotated images.
To our knowledge, this is the first dataset of its kind to be publicly released. 
The dataset is used to train a deep learning model to classify the labeled images. 
Evaluated on an independent test set, the model yields an F-1 score of 0.95, indicating that polar lows can be consistently detected from \acrshort{sar} images. 
Interpretability techniques applied to the deep learning model reveal that atmospheric fronts and cyclonic eyes are key features in the classification. 
Moreover, experimental results show that the model is accurate even if: 
\begin{enumerate*}[label=(\roman*)]
    \item such features are significantly cropped due to the limited swath width of the \acrshort{sar}, 
    \item the features are partly covered by sea ice and
    \item land  is covering significant parts of the images. 
\end{enumerate*}
By evaluating the model performance on multiple input image resolutions (pixel sizes of 500m, 1km and 2km), it is found that higher resolution yield the best performance. 
This emphasises the potential of using high resolution sensors like \acrshort{sar} for detecting polar lows, as compared to conventionally used sensors such as scatterometers. 
\end{abstract}



%% file: parts/introduction.tex
\section{Introduction}\label{sec:intro}

Polar lows belong to the class of mesoscale maritime cyclones (from now on referred to as mesocyclones) that form at high latitudes, typically due to cold air outbreaks from sea ice or snow covered regions \cite{rasmussen2003polar}. They are characterised by rapid development, small scale, strong winds and heavy snowfall. This makes them both difficult to predict and extremely hazardous for maritime activities such as fishing, shipping, petroleum extraction, and offshore wind power production. When making landfall, polar lows are prone to disrupt land and air traffic, destroy infrastructure, and trigger high snow avalanche activity in mountainous regions. 

Due to their unpredictable and destructive nature, reliable and precise methods for early detection and tracking of polar lows are desirable. Meteorologists and scientists largely rely on direct observations in terms of satellite imagery or \gls{nwp} models constrained by observations for detecting polar lows \cite{stoll2018objective, yanase2016climatology, blechschmidt20082, wilhelmsen1985climatological, zappa2014can, michel2018polar, xia2012comparison, stoll2021}. In maritime and polar regions, observations almost exclusively originate from satellites. Conventionally, data from scatterometers, radiometers and optical sensors are assimilated into the \gls{nwp} models \cite{muller2017aromearctic, hersbach2020era5}. However, these sensors either rely on sunlight or have a coarse spatial resolution (typically a few to tens of kilometres). Considering that polar lows often occur during the polar night and are small scaled, featuring wind streaks, sharp atmospheric fronts and precipitation cells, observations at higher resolution regardless of light conditions could be beneficial. 

\Glspl{sar} are independent of solar illumination and provide imagery at very high spatial resolution (typically a few to tens of metres). Researchers have already indicated that \gls{sar} data adds value to polar low monitoring \cite{moore2002, furevik2015asar, tollinger2021high}. Assimilation of \gls{sar} data into \gls{nwp} models is however challenging, since the exact relationships between radar measurement and geophysical parameters are not trivial, especially at high wind speeds \cite{tollinger2021high, chapron2001wave, mouche2012use}. An alternative approach to make use of \gls{sar} data is to rely on data driven techniques, such as deep learning.

Deep learning has successfully been applied to several remote sensing applications and achieved state of the art results \cite{zhu2017deep, bianchi2020large, bianchi2020snow, luppino2021deep}. Cyclone type phenomena specifically, has been considered in assimilated data \cite{liu2016,matsuoka2018deep, giffard2020}, passive microwave data \cite{wimmers2019using}, thermal \gls{ir} data \cite{golubkin2021, kumler2020tropical, chen2019estimating, pradhan2017tropical, krinitskiy2018deep} and scatterometer data \cite{xie2020global}. 
With the exception of \cite{carmo2021deep}, deep learning has however been largely overlooked for detecting mesocyclones in \gls{sar} data. 

This paper investigates the possibility of using deep learning for detecting mesocyclones in general, and polar lows in particular, in \gls{sar} images. 
We aim to answer two main questions: 
\begin{enumerate*}[label=(\roman*)]
    \item can a deep learning model recognise polar lows in \gls{sar} images, and 
    \item what significance  does the image resolution have on the performance? 
\end{enumerate*}

To answer these questions, we first show that a training dataset can be constructed from the Sentinel-1 data archive, which is large enough for a deep neural network to be trained. In order to make the dataset large enough, we relax the definition of a polar low to the broader class of mesocyclones. 
The constructed dataset contains image samples divided in two classes, representing the presence and the absence of mesocyclones, respectively. 
In the following, we explain in detail how the dataset is built.
To our knowledge, it is the first of its kind to be publicly released. 

Then, we show how a deep neural network can be trained on the dataset to perform binary classification with very good performance. 
The deep learning model and the training procedure is carefully motivated by considering the training dataset size, input image size, and class imbalance. 
The performance of the model is evaluated for multiple input image resolutions and interpretability techniques are applied on the model to evaluate what image features are most relevant for the classification. 









%% file: parts/data.tex
\section{SAR dataset}\label{sec:data}
This section describes the construction of the dataset for classifying mesocyclones, observed by the Sentinel-1 satellites. The dataset is publicly available (\url{https://doi.org/10.18710/FV5T9U}) and consists of \ntot{} images divided in two classes: the positive class (\npos{} images with mesocyclones) and the negative class (\nneg{} images without mesocyclones). 

\subsection{Positive class: Mesocyclone present}
To build the positive class, polar lows monitored by the Sentinel-1 satellites were required. 
Historic catalogs of polar lows exist \cite{noer2011climatological, smirnova2015polar}, based on manual analysis of \gls{nwp} model data as well as satellite data (thermal infrared, passive microwave and scatterometer data). 
However, these catalogs are regional and, more importantly, do not cover the time period when the Sentinel-1 satellites were operational.
On the other hand, studies like \cite{bracegirdle2008objective, yanase2016climatology, stoll2018objective} proposed objective criteria based on meteorological parameters that produce results similar to the manually annotated catalogs. Such objective criteria can be applied on reanalysis data, enabling identification of candidate low pressures that were coincident with the Sentinel-1 satellites. 

Although a variety of objective criteria have been proposed, they are typically associated to either: \begin{enumerate*}[label=(\roman*)]
\item the low pressure intensity, 
\item the presence of a cold air outbreak, or 
\item the location of the low pressure in relation to the polar front. 
\end{enumerate*} 
In \cite{stoll2018objective}, a combination of such criteria were imposed on the \gls{erai} dataset and the most effective criteria for detecting polar lows were identified using the manual catalog by \cite{noer2011climatological} as reference. 
However, events meeting all criteria are infrequent, since polar lows are rare. 
For reference, in \cite{noer2011climatological}, only 12 events per year were recorded over the Nordic seas on average, from year 2000 to 2009. 
Moreover, considering the limited spatio-temporal coverage of the Sentinel-1 satellites, not all events are imaged, making the number of image candidates even lower. 
Therefore, to include as many observed events as possible in our dataset, the cold air outbreak and locality type criteria were neglected. By considering only an intensity criteria, mesocyclones that are not necessarily driven by baroclinic instabilities or located in the polar air masses were included. Assuming that such mesocyclones share substantial similarities to polar lows, they can still provide valuable information to train a deep learning model, which motivates their inclusion in the dataset.

\begin{figure}[!ht]
\centering
\begin{subfigure}[b]{\linewidth}
\includegraphics[width=\textwidth]{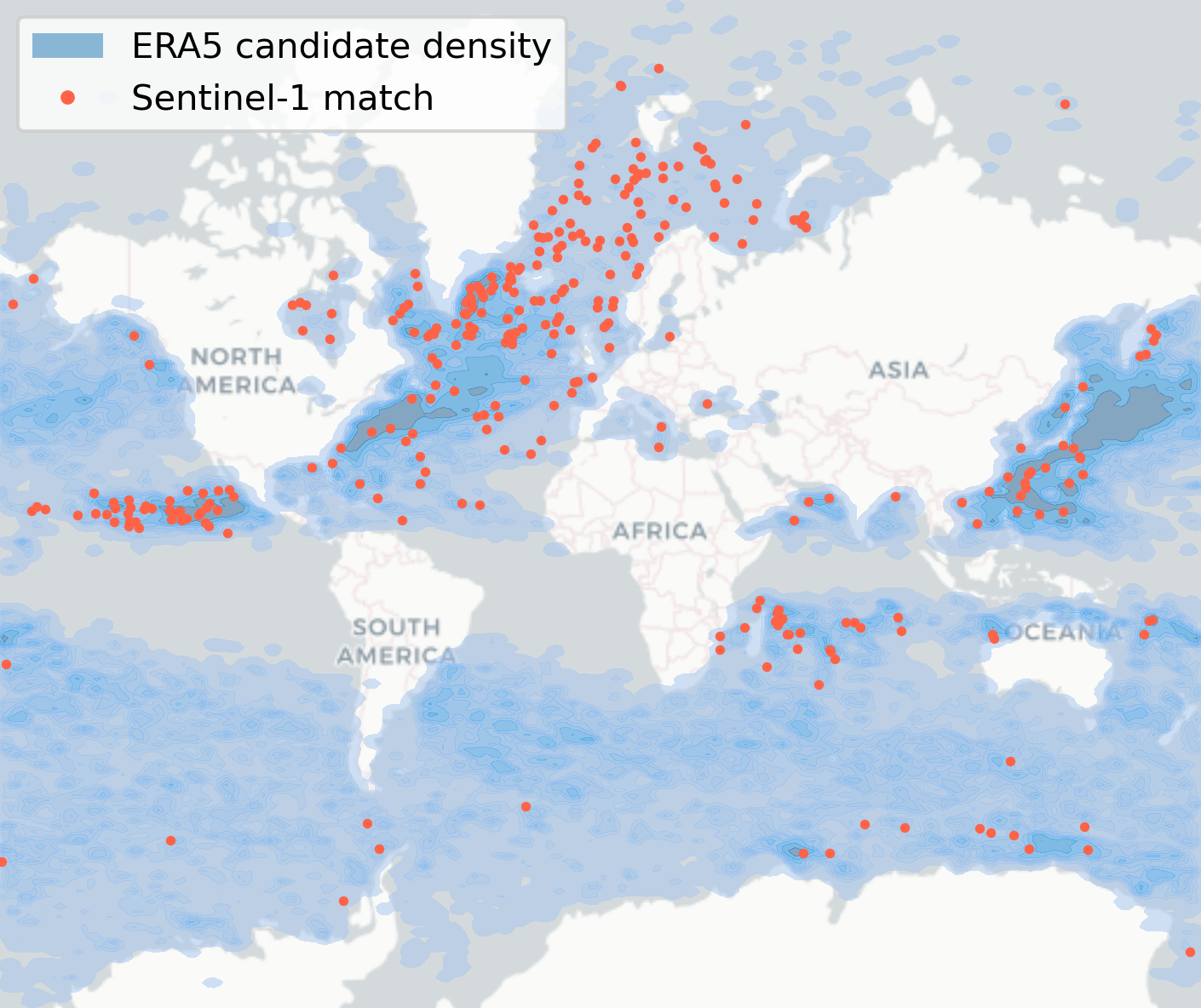}
\caption{}\label{fig:era5_s1_matches_a}
\end{subfigure}
\begin{subfigure}[b]{\linewidth}
\includegraphics[width=\textwidth]{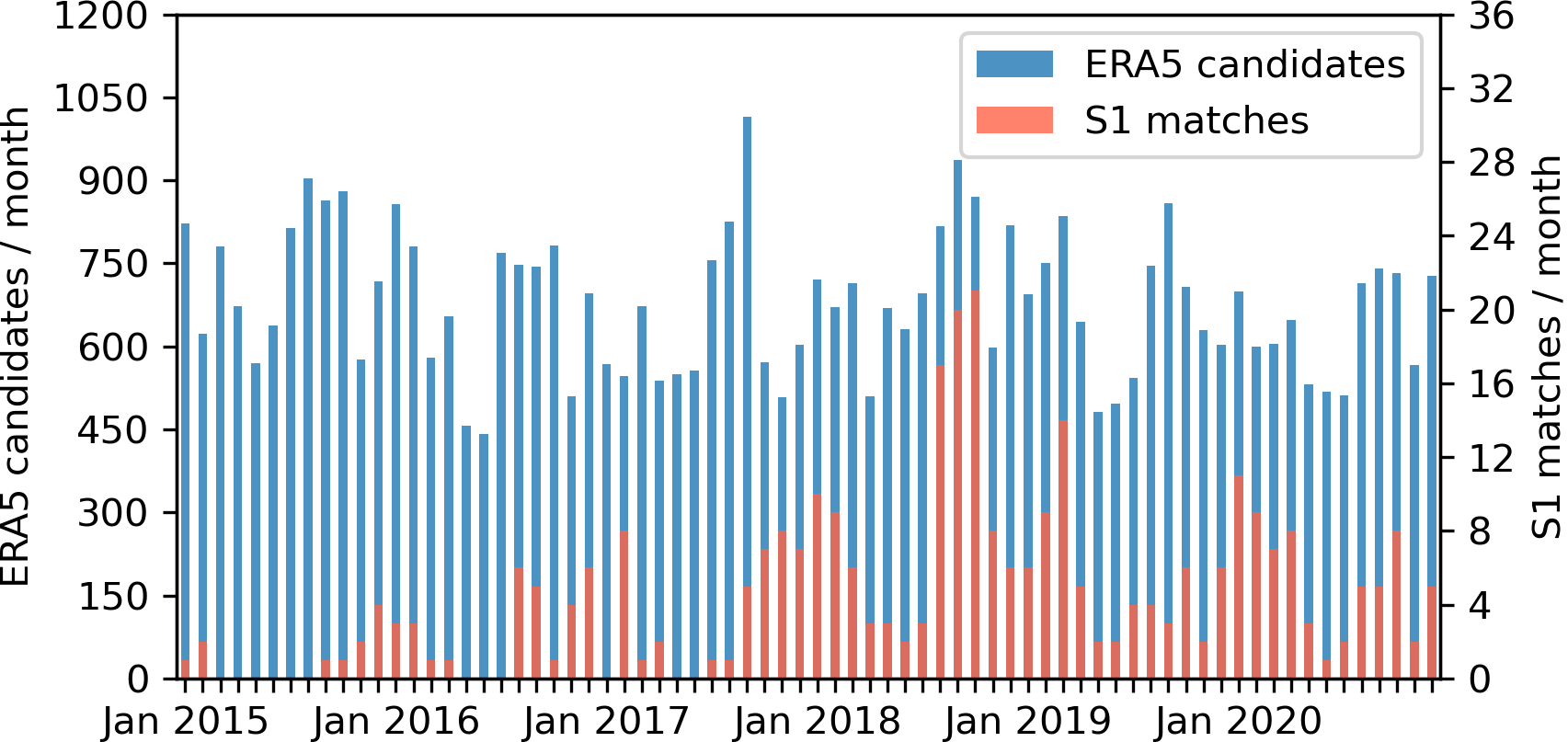}
\caption{}\label{fig:era5_s1_matches_b}
\end{subfigure}
\caption{ The spatial distribution of \gls{era5} candidates and Sentinel-1 matches in (\subref{fig:era5_s1_matches_a}). The corresponding temporal distribution, as counts per month, in (\subref{fig:era5_s1_matches_b}). Background map: \textcopyright{} OpenStreetMap contributors/CARTO. }
\label{fig:era5_s1_matches}
\end{figure}

The intensity criteria was imposed on the \gls{era5} dataset. 
Specifically, it was formulated in terms of the depression in the \gls{slp} relative to the local mean. 
This type of criteria was considered by \cite{stoll2018objective}, where different \gls{slp} depression thresholds were tested. 
In our study, the threshold was set at 230 Pa and the local mean was computed within a $9\times{}9$ grid cell neighbourhood (corresponding to $270\times{}270$ km at the equator). 
The spatio-temporal distributions of resulting candidates and the subsequent matched \gls{sar} observations are shown in figure \ref{fig:era5_s1_matches}. The highest concentrations of candidates were found in the subtropical regions of the North Pacific and North Atlantic. However, due to higher satellite revisit frequencies at higher latitudes, most \gls{sar} observations were found in the extra tropical and polar parts of the North Atlantic. 

\begin{figure}[ht!]
\centering
\includegraphics[width=.95\linewidth]{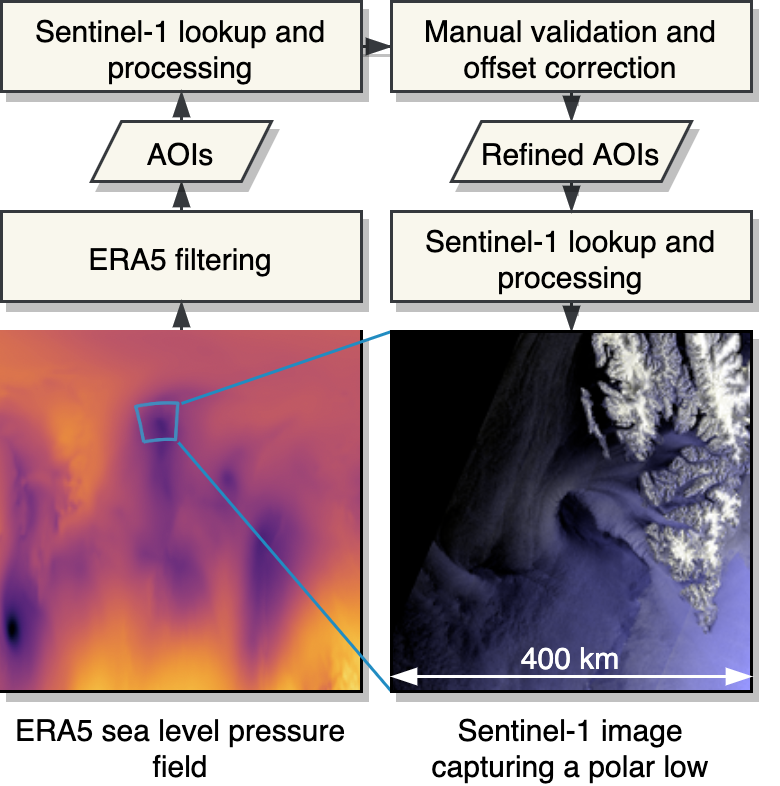}
\caption{ The dataset construction process illustrated as a flow chart. The starting point is the \acrshort{era5} \acrshort{slp} field that is filtered according to our working definition of a polar low to form candidate \acrshort{aoi}s. These are used to search for Sentinel-1 data, which is processed (see section \ref{sec:sarprocessing}) onto a \acrshort{utm}-grid. The processed images are validated manually and offset corrected by refining the AOIs and reprocessing the images onto a \acrshort{utm}-grid in which the image features are centered. The polarisation channels for this image sample is HH/HV.}
\label{fig:dataset_making_flowchart}
\end{figure}

The dataset construction process is illustrated in figure \ref{fig:dataset_making_flowchart} and each step is described in detail below.

\subsubsection{ERA5 filtering}
The \gls{era5} dataset \cite{hersbach2020era5} consists of hourly reconstructions of a large number of meteorological variables, spanning from 1950 to present. The data can be accessed on a geodetic grid, with a grid spacing of 0.25 degrees horizontally. Considering the global grid, candidate low pressures were identified by: \begin{enumerate*}[label=(\roman*)]
\item low-pass filtering the \gls{slp} using a $9{\times}9$ sliding average filter, 
\item selecting candidate grid cells where the \gls{slp} was 230 Pa lower than the low-pass \gls{slp}, 
\item grouping adjacent candidate grid cells and
\item keeping groups with an equivalent radius smaller than 200 km (i.e.\ with an area less than $200^2\pi$ km${}^2$, thus excluding very big weather systems).
\end{enumerate*} Each such group was vectorised and constituted a candidate \gls{aoi}. This filtering process was done from 1 Januray 2015 to 31 December 2020 with a time step of 3 hours. 

\subsubsection{Sentinel-1 lookup and processing}\label{sec:sarprocessing}
For each candidate \gls{aoi} at time $t$, we queried the Copernicus Open Access Hub for Sentinel-1 \gls{grd} products in the time interval $t \pm 1.5$ hours.
Resulting products were downloaded from \gls{asf} and processed\footnote{
All \gls{sar} data processing was done using \gls{gdar}, a python based library for processing raster data in radar geometries, developed by \gls{norce}} by: 
\begin{enumerate*}[label=(\roman*)]
\item calibrating the data to sigma-nought, 
\item removing thermal noise, 
\item merging time-adjacent products to a common grid in \gls{sar} geometry,
\item multi-looking to 500 m resolution in range and azimuth\footnote{In terms of number of looks, \acrshort{ew} mode products are in total multi-looked by $60\times20$ looks in range and azimuth, while \acrshort{iw} mode products are multi-looked by $250\times50$ looks in range and azimuth. Speckle noise is thus significantly suppressed in the processed images. }, 
\item geocoding to a 400$\times$400 km grid (centered at the \gls{aoi}) in a \gls{utm} coordinate system with a 500 m grid spacing and 
\item generating \gls{rgb} colour composites. 
\end{enumerate*}

The \gls{rgb} colour composites were generated by first re-scaling the radar cross-section (assumed in decibel scale) to a value $x \in [0,1]$. Specifically, the 2$^\textrm{nd}$ and 98$^\textrm{th}$ percentiles of each separate image and polarisation channel were re-scaled to 0 and 1, respectively\footnote{
The 2$^\textrm{nd}$ percentile was clipped to the range -25 to -15 dB and the 98$^\textrm{th}$ percentile was clipped to the range -10 to 0 dB. The clipping values were chosen to harmonise the scaling across image samples. Pixels without data were excluded when computing the percentiles and replaced by zeros. }.

For data with dual polarisation channels, the re-scaled values were used to make \gls{rgb} colour composites as: 
\begin{equation*}
\textrm{R} = \textrm{G} = \frac{x_{\rvert\rvert} + x_{\times}}{2} ~,~~
\textrm{B} = x_{\rvert\rvert}
\end{equation*}
where $x_{\rvert\rvert}$ and $x_{\times}$ corresponds to the co- and cross-polarised\footnote{
The co-polarised channel can be either HH or VV, and the corresponding cross-polarised channel can be either HV or VH.} channels, respectively. For single polarisation data, containing only the co-polarised channel, the colour channels were defined as: $\textrm{R} = \textrm{G} = \textrm{B} = x_{\rvert\rvert}$. Both dual and single polarisation data were thus considered jointly in the training data set\footnote{A dedicated experiment using only the co- or cross-polarised channel separately, can be found in the supplementary material. }, however, the dual polarisation data constituted the great majority of the samples (see figure \ref{fig:set_histograms}).

\subsubsection{Manual validation and offset correction}
Each \gls{rgb} colour composite was manually validated. Specifically, in each positive image, we asserted the presence of distinctive features (typically an eye or a comma shaped pattern). 
In general, these features were not centered in the processed images, since the image grid was centered at the candidate \gls{aoi} originating from \gls{era5}. Therefore, offsets were corrected for by manually centering the \glspl{aoi} on the eye or comma shaped pattern. The samples were then reprocessed with the refined \glspl{aoi}.


\begin{figure*}[!ht]
\centering
\begin{subfigure}[b]{0.496\textwidth}
\includegraphics[height=\textwidth]{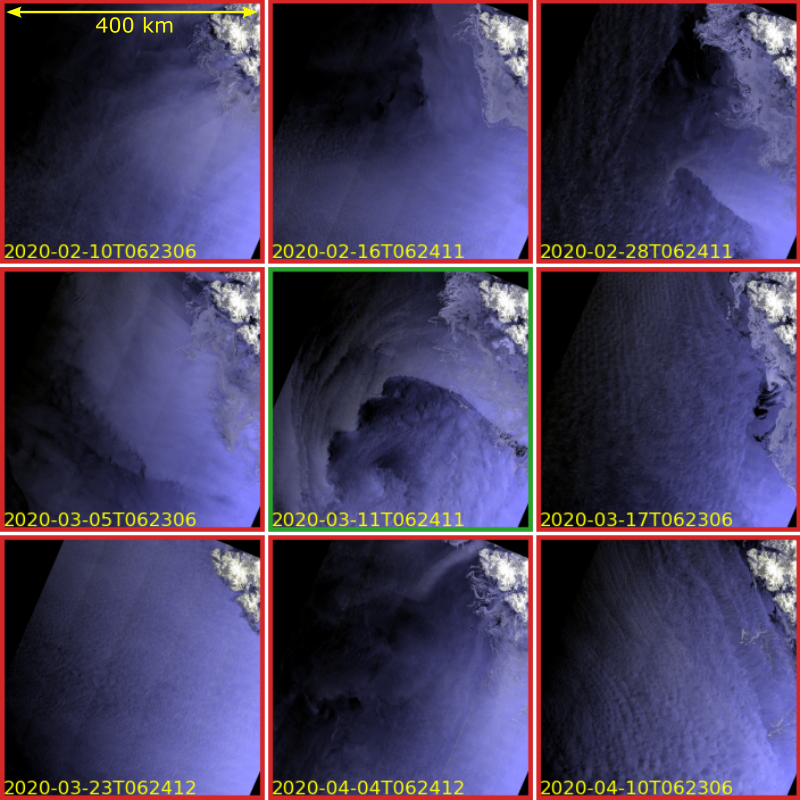}
\caption{}\label{fig:s1_pos_neg_set_a}
\end{subfigure}
\begin{subfigure}[b]{0.496\textwidth}
\includegraphics[height=\textwidth]{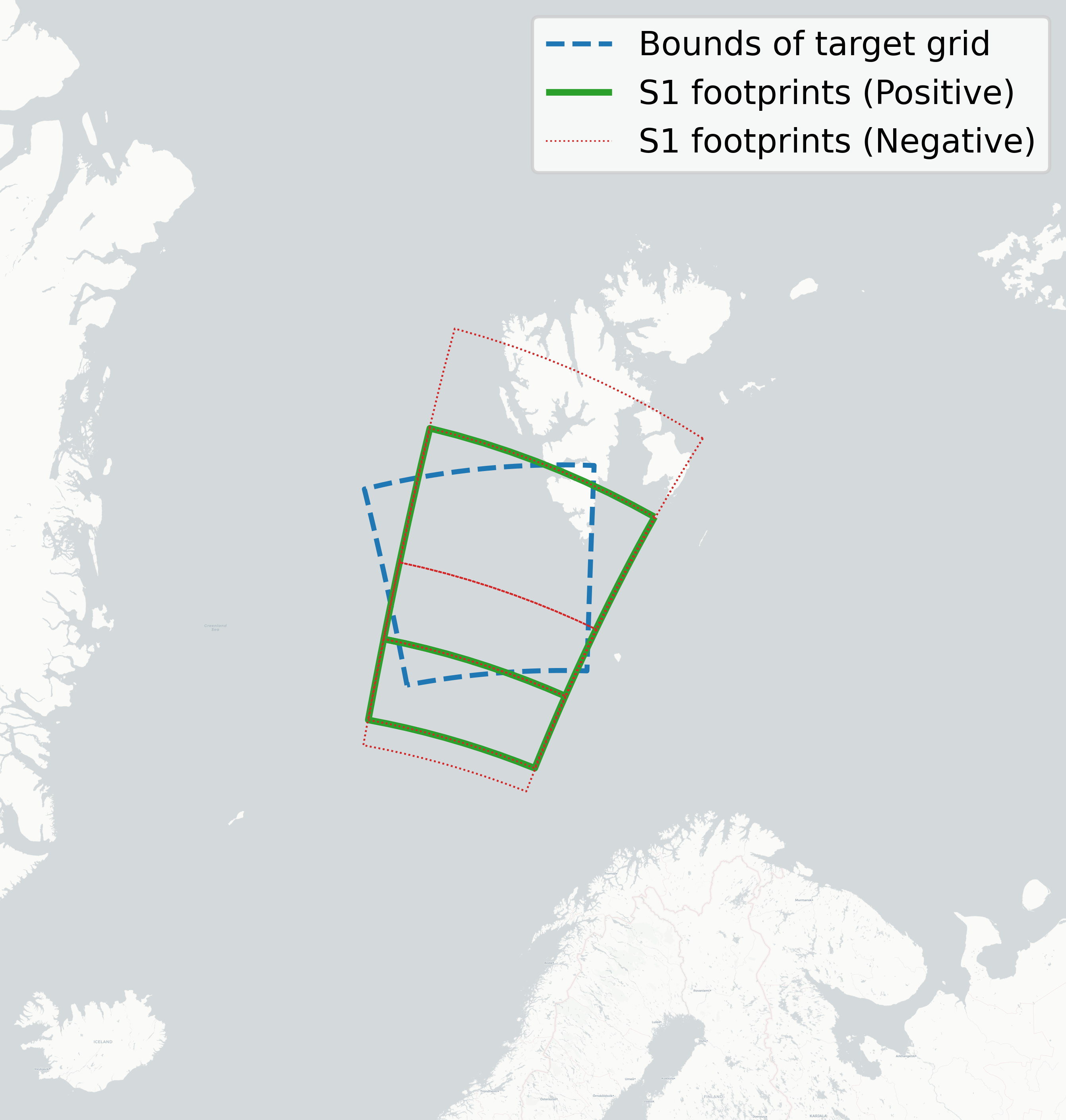}
\caption{}\label{fig:s1_pos_neg_set_b}
\end{subfigure}
\caption{ In (\subref{fig:s1_pos_neg_set_a}), a repeat-pass set is shown, consisting of one positive (central image) and eight negative samples. All samples within a set are processed onto the exact same grid, centered at the low pressure in the positive sample. The polarisation channels for this set are HH/HV. In (\subref{fig:s1_pos_neg_set_b}), the location of the individual products and the grid is displayed. Background map: \textcopyright{} OpenStreetMap contributors/CARTO. }
\label{fig:s1_pos_neg_set}
\end{figure*}

\subsection{Negative class: Mesocyclone absent}\label{sec:negatives}
To obtain samples of the negative class, representing the absence of a cyclone, we considered repeat-pass \gls{sar} acquisitions successive to those of the positive image samples (i.e.\ images acquired at the same relative orbit). 

The motivation of our choice was twofold: 
\begin{enumerate*}[label=(\roman*)]
\item Sentinel-1 repeat-passes are separated by at least 6 days, which is enough time for the sea state (and thus the image features) to decorrelate, and  
\item the imaging geometry of repeat-pass acquisitions is nearly identical, such that static/background features appear similar. 
\end{enumerate*}
The second point is important in order to factor out land features from the dataset. Indeed, if the same land features appear in both the positive and negative class, it is expected that a machine learning model will be able to ignore them in the classification task. 

As an example, a repeat-pass image set consisting of one positive and eight negative samples is shown in figure \ref{fig:s1_pos_neg_set}. To the left, the processed \gls{rgb} composites are shown. The south tip of Svalbard can be seen statically in all images, while ocean features appear dynamically. The positive sample in the centre, contains a distinct vortex structure. To the right, a map with the footprints of the individual Sentinel-1 products involved is shown, together with the footprint of the image grid. Typically, due to the limited swath of the \gls{sar}, the products do not cover the whole image grid across track, leading to missing data in the \glspl{rgb} in the cross-track direction. Occasionally, some products are not captured, leading to missing data in the along-track direction as well. 

\begin{figure}[!ht]

\centering
\includegraphics[width=.8\linewidth]{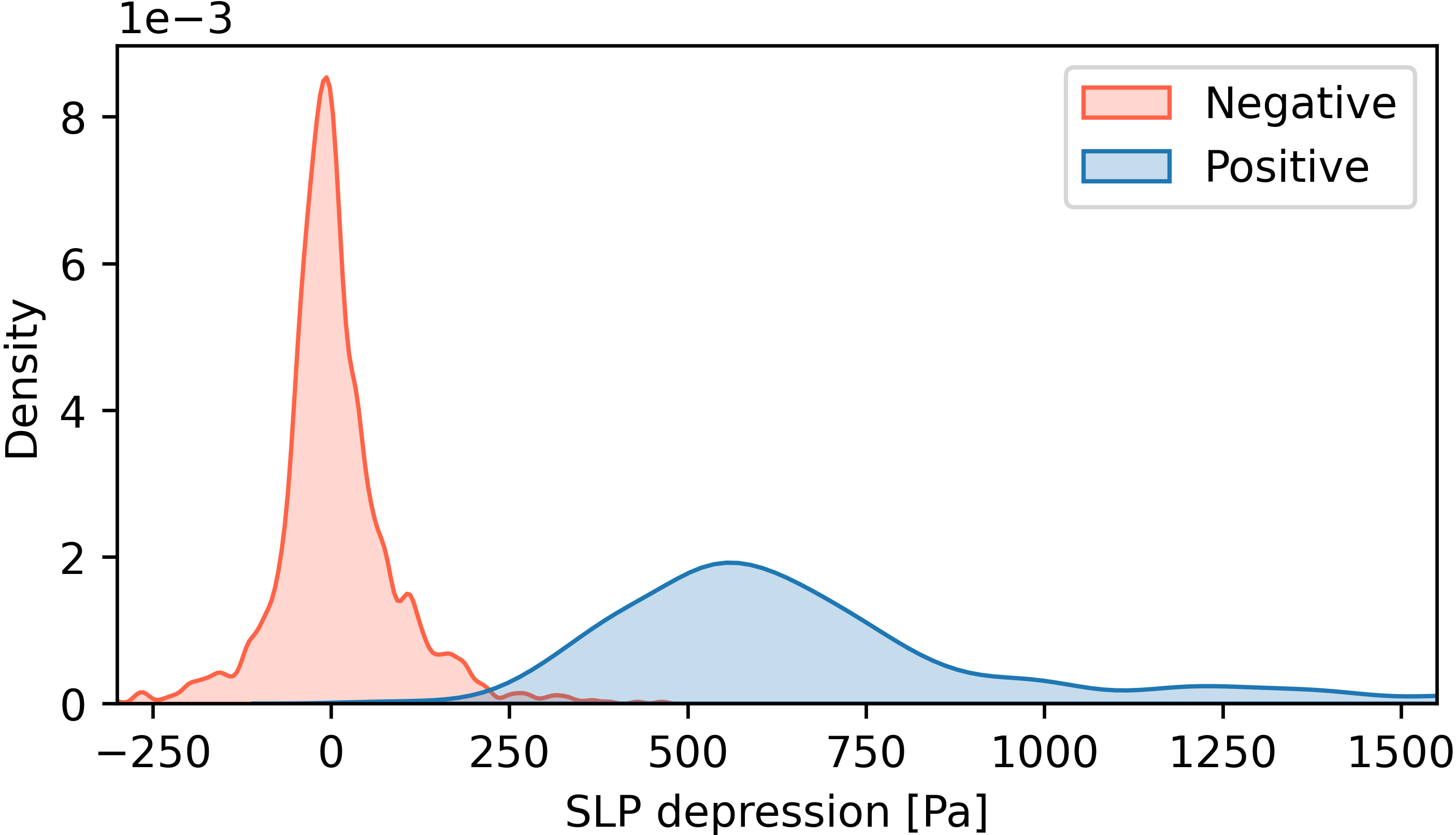}
\caption{ The distribution of \gls{slp} depression for the two classes. The depression is measured as the \gls{slp} averaged over the whole image, minus the \gls{slp} averaged over the centre $100{\times}100$ pixels. }
\label{fig:set_slp_densit}
~\\
\includegraphics[width=.8\linewidth]{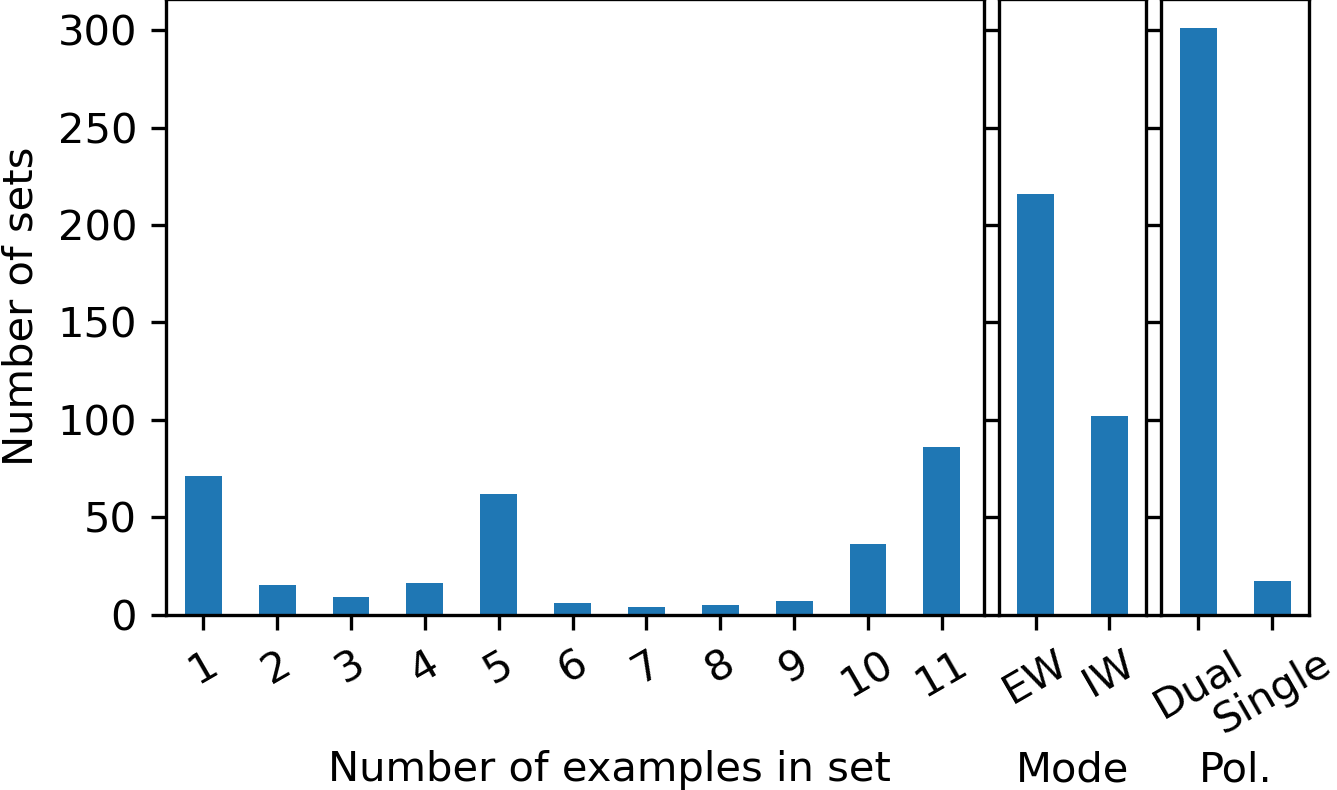}
\caption{ Overview of the repeat-pass sets, in terms of distributions in size (samples per set), imaging mode and polarisation mode. For each positive sample, we extracted a maximum of 10 negative samples.}
\label{fig:set_histograms}
\vspace*{-15pt}
\end{figure}

The distribution of \gls{slp} depression, defined as the difference in \gls{slp} (extracted from \gls{era5}) between the image wide average and the average of the centre 100$\times$100 pixels, is shown in figure \ref{fig:set_slp_densit}. The positive samples have a strong depression, while the negative samples exhibit a symmetric distribution centred around 0 Pa. This indicate that the negative class, indeed, represents a sea state not biased towards a centered low pressure. It should however be emphasised that no particular features are excluded from the negative class. 

In total, \nneg{} negative sample were generated from the \npos{} positive ones, resulting in a total of \ntot{} samples in the dataset. The number of negatives per positive varied depending on the existence of repeat-pass acquisitions, as shown in figure \ref{fig:set_histograms}. Furthermore, most samples were acquired in the \gls{ew} mode with two polarisations.

%% file: parts/machine_learning.tex
\section{Deep learning}\label{sec:ml}

Three immediate challenges can be identified when choosing and training an appropriate deep learning model to perform classification on the dataset: 
\begin{enumerate*}[label=(\roman*)]
    \item the input image size is relatively large, 
    \item the training dataset is relatively small, and
    \item the classes are imbalanced. 
\end{enumerate*}
In the following, we discuss how these were dealt with. 

\subsection{Deep learning architecture}\label{sec:architecture}
One of the major benefits of using \gls{sar} data, compared to e.g.\ scatterometer or passive microwave data, is the high image resolution. Although the images in the training dataset were already heavily downsampled from the original resolution of 10-40 m to 500 m, resulting in an image size of $800\times800$ pixels, they are relatively large in the context of many popular deep learning models. These are often designed for images of size $256\times256$ pixels or lower. To preserve details that are specific for the \gls{sar} data, such as wind streaks, rain cells or sharp atmospheric fronts, and to enable us to study the added value of high input image resolution, we wish to avoid further downsampling and rather let the model handle the high input image resolution. 
\Glspl{cnn} used for image classification usually consist of a stack of convolutional layers followed by pooling. Each such processing block sequentially increases the feature dimensionality through colvolutions, while reducing spatial resolution through pooling. As such, relevant spatial information will gradually become embedded in the feature space. If the input image is large, the model must either apply an aggressive downsampling in each processing block, or include many blocks and, thus, become very deep. The former can be obtained by large stride in the convolutional and pooling layers, or by using Atrous convolutions~\cite{chen2017deeplab}. These techniques do, however, come at a cost of discarding spatial information, which we wish to avoid. This leaves us with the option of using a deep architecture, which gradually distill the spatial information and embeds it into the feature space.

Training a very deep network poses two fundamental challenges. Firstly, the gradients of the loss used to update the parameters may gradually vanish as they are backpropagated through the network. Secondly, an architecture with many layers contain many trainable parameters. This makes the model prone to overfitting, unless the training set is exceptionally large, which was not the case in our study. 

A solution to address the first problem is to use residual connections, popularized by architectures such as ResNet~\cite{szegedy2017inception}, which facilitate the flow of the gradients during the backpropagation.

Considering the second problem, a ResNet is unfortunately characterized by many trainable parameters. There are, however, more recent deep architectures which include residual connections but have fewer parameters. For example, MobileNet~\cite{howard2017mobilenets} and Xception~\cite{chollet2017xception} implement \gls{sep2d}, which allows to greatly reduce the number of trainable parameters\footnote{The basic idea behind a \gls{sep2d} is to replace a matrix of parameters $\mathbf{W} \in \mathbb{R}^{N\times M}$ with an outer product of two unidimensional vectors, $\mathbf{\tilde W} = \mathbf{u}^T \otimes \mathbf{v}$, where $\mathbf{u} \in \mathbb{R}^M$ and $\mathbf{v} \in \mathbb{R}^N$, reducing the number of parameters from $M \cdot N$ to $M + N$. This is, actually, a simplification. In practice, a \gls{sep2d} splits the traditional convolution with a kernel of size $H \times W \times F_\text{in} \times F_\text{out}$ with a depth-wise convolution with $F_\text{in}$ kernels of size $H \times W \times 1$, followed by a point-wise convolution with a kernel of size $1 \times 1 \times F_\text{out}$.}. 

\begin{figure}[!ht]
    \centering
    \includegraphics[width=.7\columnwidth]{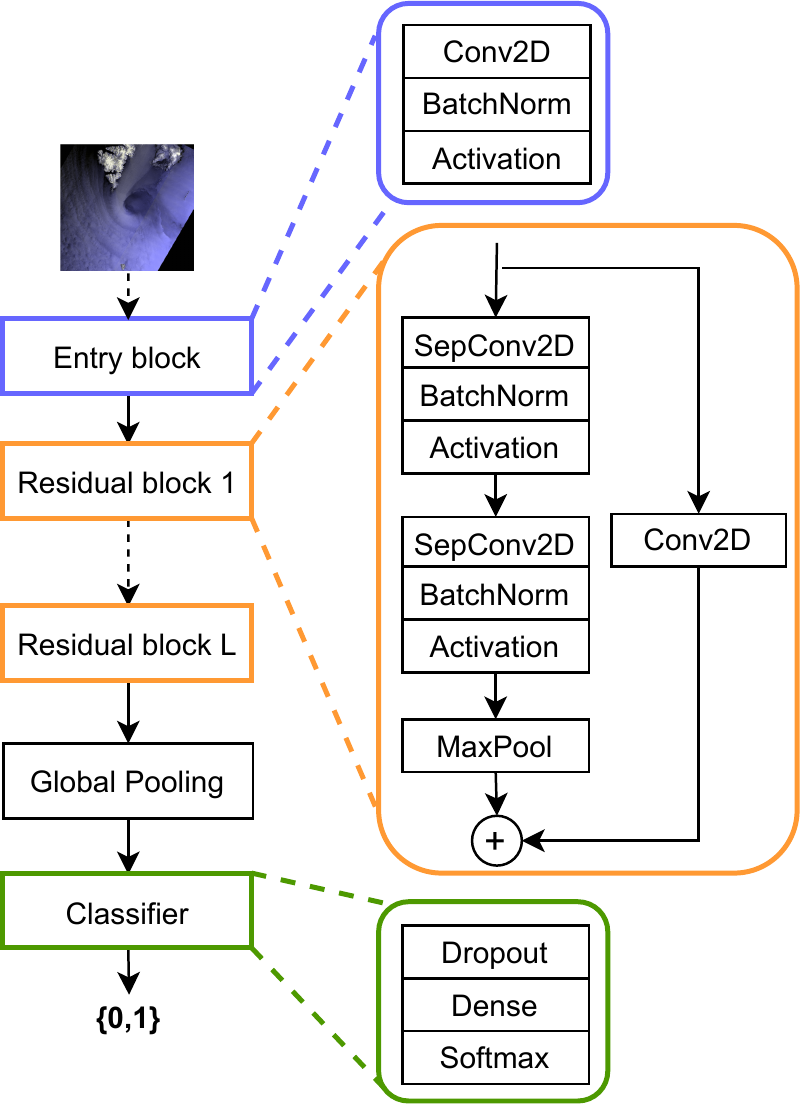}
    \caption{ Architecture details.}
    \label{fig:architecture}
\end{figure}

Therefore, we opted for a customized Xception architecture\footnote{A comparison with other popular deep learning architectures is presented in the supplementary material.}, whose details are depicted in figure \ref{fig:architecture}.
The entry block consists of a convolutional layer, followed by a batch normalization layer~\cite{ioffe2015batch} and a ReLU activation function.
There are $L$ residual blocks, each one including \gls{sep2d} layers, batch normalization, ReLU activations and a max-pooling layer.
The max-pooling output is combined with the input of the residual block through a skip connection.
The convolutional layer in the middle of the skip connection has no activation function and simply applies a kernel of size 1 with stride 2, to match the shape of the input with the one of the output.
A global pooling layer reduces the feature map generated by the last residual layer to a single vector, which is processed by the final classifier consisting of a dropout layer~\cite{srivastava2014dropout}, a dense layer, and a softmax activation.

\subsection{Data augmentation}\label{sec:aug}
Augmenting the training data by applying random transformation is a common technique used to prevent overfitting. By exposing the deep learning model to perturbations of original inputs, it is possible to improve the robustness of the model.
In addition, data augmentation allows to get rid of some bias in the dataset and increase the generalization performance on new unseen data.

\begin{figure}[!ht]
    \centering
    \includegraphics[width=\columnwidth]{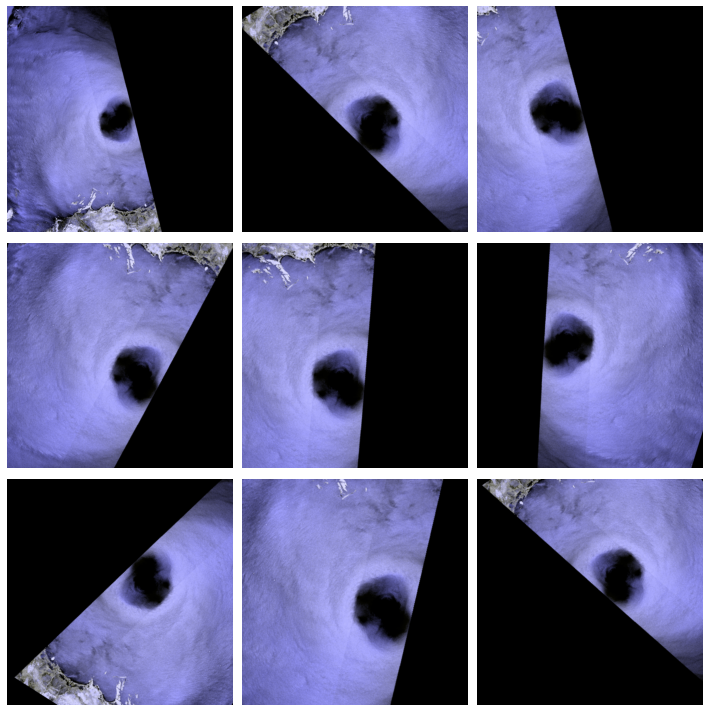}
    \caption{ Examples of random image augmentation. The top-left image is the original. The polarisation channels for the image are VV/VH. }
    \label{fig:augmentation}
\end{figure}

Our dataset has been designed by keeping image augmentation in mind.
Each low pressure is centered in the image and has a wide area around that can be partially cropped.
Each time a batch of images is fetched to our deep learning model, the following random transformations are applied on the fly; 
\begin{enumerate*}[label=(\roman*)]
    \item horizontal and vertical translation (between 0 and 10\% of the image size), 
    \item horizontal and vertical flip, 
    \item rotation (0 to 40 degrees), 
    \item zoom (-10\% to 10\% of the original scale) and finally
    \item cropping to the centre $512 \times 512$ pixels. 
\end{enumerate*}
If after the transformation some points fall outside the boundaries of the original input image, these are filled with zeros.
Notably, after data augmentation the low pressures are no longer centered in each image. Figure \ref{fig:augmentation} shows an example of augmented images randomly generated during training.

\subsection{Class imbalance}
While the number of positive samples were restricted by the number of matches found on the Sentinel-1 archive, multiple negative samples could be generated for each positive sample. 
This lead to a natural skewness in the distribution of classes in the dataset: 84\% of the samples belong to the negative class and 16\% to the positive.
We tested and compared three different approaches to train the deep learning model in the presence of class imbalance. 

\subsubsection{Class-weighting}
We re-weighted the loss function according to class frequencies. 
Denoting the number of samples in the negative and positive classes respectively by $n_0$ and $n_1$, the loss for samples in the corresponding classes were weighted by $w_0 = \frac{1}{n_0} \frac{n_0 + n_1}{2}$ and $w_1= \frac{1}{n_1} \frac{n_0 + n_1}{2}$. 
This means that each error on the positive class affects the optimization of the model weights to a greater extent.

\subsubsection{Oversampling}
We ensured that in each batch there were always the same amount of samples of both classes. Specifically, this was obtained by designating half of the batch to the negative class, and half to the positive. In each epoch\footnote{One epoch is when the whole training dataset has been passed forward and backward through the network once. }, samples from the negative class are seen only once, while samples from the positive class are repeated. 
We made sure to observe all samples of the negative class at least once during an epoch and to not sample any image twice within a batch. 

\subsubsection{Rejection sampling} 
This strategy drops samples until a balanced distribution across the two classes is obtained.
Contrarily to oversampling, each sample is seen at most once in each epoch, which makes the overall training faster.

By empirical comparison (see the supplementary material), we found that oversampling yields the best performance and, therefore, was the strategy adopted in our experiments.

\subsection{Hyperparameter tuning}
To find the optimal configuration of the deep learning model, we searched several hyperparameters and selected those giving the best performance on a validation set.
As validation set, we used 10\% of the training set.
The hyperparameter space and the optimal values found after the optimization procedure are reported in table \ref{tab:hyperparams}.
To reduce the hyperparamers space, we only search the number of filters of the first residual blocks and then we double the number in the following blocks.

\begin{table}[!ht]
\footnotesize
\centering
\begin{tabular}{lll}
\hline
\textbf{Hyperparam.}                       & \textbf{Search space}          & \textbf{Optimal} \\ \hline
Activation                        & \{ReLU, SeLU\}        & ReLU    \\
Conv2D filters                    & \{8, 16\}             & 8       \\
Kernel size                       & \{3, 5\}              & 3       \\
SepConv2D filters                 & \{8, 16, 24, 32\}     & 8       \\
Num. residual blocks              & {[}2, 8{]}            & 7       \\
Global pooling                    & \{avg, flat, max\}    & avg     \\
Num. dense layers                 & {[}1, 3{]}            & 1       \\
Units in the dense layer          & \{8, 16, 24, 32\}     & 8       \\
Dropout rate                      & {[}0.1, 0.6{]}        & 0.5     \\
Use batch normalization           & \{True, False\}       & True    \\
Learning rate                     & \{1e-2, 1e-3, 1e-4\}  & 1e-3    \\ \hline
\end{tabular}
\caption{ Hyperparameters space and optimal values found. ``Conv2D filters'' and ``kernel size'' refer to the entry block. ``SepConv2D filters'' refers to the 1\textsuperscript{st}  residual block, since the number of filters is double each time in the following blocks.}
\label{tab:hyperparams}
\end{table}

Since the dataset contains large images and we consider deep models with many parameters, evaluating each hyperparameter configuration is computationally expensive.
Therefore, rather than performing an exhaustive search with grid search or evaluating a large number of configurations with a random search, we opted for a more efficient approach.
In particular, we used Bayesian hyperparameter optimization \cite{snoek2012practical}. 


We used a batch size of 16 and the Adam optimizer~\cite{kingma2014adam}.
During the hyperparameter tuning we trained the model for 50 epochs.
After finding the optimal configuration, we trained the final model for 200 epochs.

\subsection{Model interpretability}\label{sec:interp}
Due to the presence of many non-linear transformations, it is difficult to interpret the decision process of a neural network and considerable research effort has been devoted to improve our understandings.
Gradient based approaches try to find which inputs have the most influence on the model scoring function for a given class~\cite{smilkov2017smoothgrad, zeiler2014visualizing, springenberg2014striving}.
This is usually done by taking the gradient of the class activation score  with respect to each input features~\cite{simonyan2013deep}.
A drawback of gradient based methods is that they give zero contribution to inputs that saturate the ReLU or MaxPool.
To capture such shortcomings, a formal notion of explainability was introduced in \cite{bach2015pixel} with the axiom of conservation of total relevance, which states that the sum of relevance of all pixels must match the class score of the model.
Specifically, the authors propose to distribute the total relevance of the class score to the input features with Layer-wise Relevance Propagation (LRP).
While LRP follows the conservation axiom, it does not specify how to distribute the relevance among the input features. To address this problem DeepLiFT~\cite{shrikumar2017learning} enforces an additional axiom on how to propagate the relevance by following the chain rule.

In this work, we adopt two recent interpretability techniques, that address some of the shortcomings discussed above and are able to provide valuable insights into the decision problem of our model.

\subsubsection{Integrated Gradients}
\Gls{ig}~\cite{sundararajan2017axiomatic} has become a popular interpretability technique since it can be applied to any neural network model, is easy to implement, and theoretically grounded.
\Gls{ig} aims to satisfy two additional axioms that are not jointly ensured by other existing attribution schemes; 
\begin{enumerate*}[label=(\roman*)]
    \item if the input and an uninformative baseline differ in exactly one feature, such a feature should be given non-zero attribution, 
    \item when two models are functionally equivalent, they must have identical attributions to input features. 
\end{enumerate*}

Denoting the model scoring function $F$, the attributions given by \gls{ig} are
\begin{equation}
    \label{eq:ig}
    \text{IG}(x) := (x - x') \cdot \int_{\alpha=0}^1 \frac{\partial F \left( x' + \alpha \cdot (x -x') \right)}{ \partial x} d\alpha,
\end{equation}
where $x$ is a sample in the dataset, $x'$ is the uninformative baseline, and $\alpha$ is an interpolation constant used to perturb the input features.

\begin{figure}[ht!]
    \centering
    \includegraphics[width=\columnwidth]{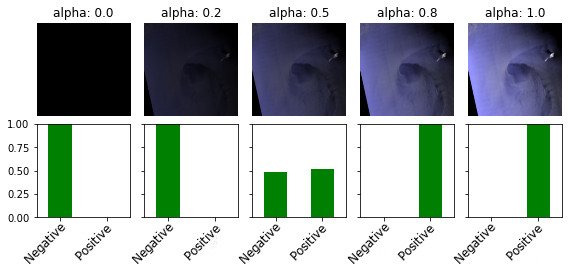}
    \caption{ Top row: linear interpolation from the zero-baseline (left) to an actual sample of positive class (right). 
    Bottom row: classification probabilities assigned by the model at each step of the interpolation.}
    \label{fig:interp}
\end{figure}

In our study, we let $x'$ be a black image (all zeros) as the uninformative baseline.
As empirically confirmed in our experiments, such a baseline is classified with high confidence to be negative.
Let $\mathcal{X}$ be the set of interpolated images from $x'$ to $x$.
The computation of the integral in \eqref{eq:ig} is approximated with the sum of the partial derivatives of the images in $\mathcal{X}$. 
%
%
Figure \ref{fig:interp} depicts a small interpolation set $\mathcal{X}$ from the mean-baseline to a positive sample and shows how the classification score changes.
By summing the gradients $\frac{\partial F(\mathcal{X})}{\partial x_i}$ one quantifies the relationship between the changes in the input features and the changes in the predictions of the model $F$.

\subsubsection{Gradient-weighted Class Activation Mapping}\label{sec:gcam}
While \gls{ig} can be used on any neural network model, \gls{gcam} is specific for \glspl{cnn}. It uses the gradients of a given target class flowing into the final convolutional layer to produce a coarse localization map, which highlights the important regions in the image for predicting the class~\cite{selvaraju2017grad}.
We summarize at a high level the main steps of the algorithmic procedure and we refer the interested reader to \cite{selvaraju2017grad} for further details: 
\begin{enumerate*}[label=(\roman*)]
    \item take a trained model and cut it at the $k$-th layer, which is the layer for which we want to create a Grad-CAM heat-map (usually, the activation after the last convolutional layer), 
    \item feed an input image to the complete model and collect the total loss and the output of layer $k$, 
    \item compute the gradients of the output of layer $k$ with respect to the loss, 
    \item take parts of the gradient which contribute to the prediction and use to build a heatmap, and 
    \item resize the heatmap so that it can be overlaid to the original image.
\end{enumerate*}

%% file: parts/results.tex
\section{Results}\label{sec:results}
The dataset presented in section \ref{sec:data} was used to train the model described in section \ref{sec:ml}. Specifically, the dataset was partitioned such that 79\% of the samples were used for training and validation and 21\% for testing. The partitioning was done by randomly assigning complete repeat-pass sets to either the test or the training set. In such a way, positive and negative samples with the same land features cannot appear both in the training and test set. This, 
\begin{enumerate*}[label=(\roman*)]
    \item encouraged the model to factor land features out as irrelevant to the classification task, 
    \item allowed us to evaluate the generalization capability of the model by testing on new locations, unseen during training.
\end{enumerate*} 

The arguably most attractive property of \gls{sar} data, as compared to e.g.\ scatterometer data, is the high spatial resolution. In order to evaluate the added value of higher spatial resolution, the model accuracy was examined for three different input image resolutions\footnote{The highest resolution here (500 m) is still considerably lower than the original resolution of the \gls{sar} images. However, as discussed in section \ref{sec:architecture}, the input image size is limited by the depth of the network architecture in relation to the size of the training dataset. Therefore, we did not considered even higher input image resolutions, even if the original data allowed for it. }; 500 m, 1000 m and 2000 m (the latter two obtained by bi-linear down sampling of the first).  
Hyperparameter tuning was performed independently for each resolution (see the supplementary material for details), and the classification performance on the test set is shown in table \ref{tab:res_all_folds}. The table displays the mean and standard deviation of \glspl{tn}, \glspl{fn}, \glspl{fp}, \glspl{tp} and F1 score obtained from 10 independent runs. It is clear that higher image resolution significantly improves the classification results\footnote{A detailed comparison based on \gls{gcam} between the model trained on 2000 m and 500 m resolution is presented in the supplementary materials. }. In fact, for the highest input resolution, the model is misclassifying on average less than 8 samples (as \gls{fn} or \gls{fp}) out of the \ntest{} samples in the test set, with a mean F1 score of 0.94 (in the supplementary materials, results of the performance for the highest input image resolution using the co- or cross-polarised channels separately are also presented). 
\begin{table}[!ht]
\centering
\footnotesize	
\begin{tabular}{c|ccccc}
Pixel size & \gls{tn} & \gls{fn} & \gls{fp} & \gls{tp} & F1 score \\ \hline
2km & 346.6{\tiny$\pm$2.1} & 6.4{\tiny$\pm$2.1} & 9.8{\tiny$\pm$1.9} & 54.2{\tiny$\pm$1.9} & 0.87{\tiny$\pm$0.01} \\ %
1km & 364.4{\tiny$\pm$2.1} & 6.6{\tiny$\pm$2.1} & 7.4{\tiny$\pm$2.1} & 56.6{\tiny$\pm$2.1} & 0.89{\tiny$\pm$0.02} \\ %
500m & 367.8{\tiny$\pm$2.7} & 3.2{\tiny$\pm$2.7} & 4.6{\tiny$\pm$1.7} & 59.4{\tiny$\pm$1.7} & 0.94{\tiny$\pm$0.01} \\ %
\hline
\end{tabular}
\caption{ Classification performance on the test set when using different input resolutions. It is evident that higher input image resolution significantly improves the performance. }
\label{tab:res_all_folds}
\end{table} 

\begin{table}[!ht]
\centering
\begin{tabular}{ccccc}
TN & FN & FP & TP & F1 score \\ \hline
366 & 2 & 5 & 62 & 0.95 \\
\hline
\end{tabular}
\caption{ Classification performance obtained on the specific run where we apply the interpretability techniques.}
\label{tab:res_specific_fold}
\end{table}

A model trained on the 500 m resolution images was further examined using the \gls{ig} and \gls{gcam} techniques presented in section \ref{sec:interp}. The performance of this specific model is shown in table \ref{tab:res_specific_fold} and the images it classifies as \glspl{tp}, \glspl{fp}, and \glspl{fn} are discussed in the following. 
The deep learning model used in our experiments and the code to apply the interpretability techniques is available online\footnote{ \url{https://github.com/FilippoMB/Recognition-of-polar-lows-in-Sentinel-1-SAR-images-with-deep-learning}}.

\begin{figure*}[ht!]
\captionsetup[subfigure]{labelformat=empty}
	\centering
        \includegraphics[width=0.2\textwidth]{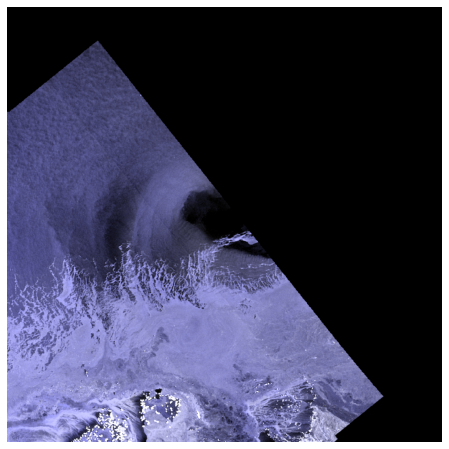}
        \includegraphics[width=0.2\textwidth]{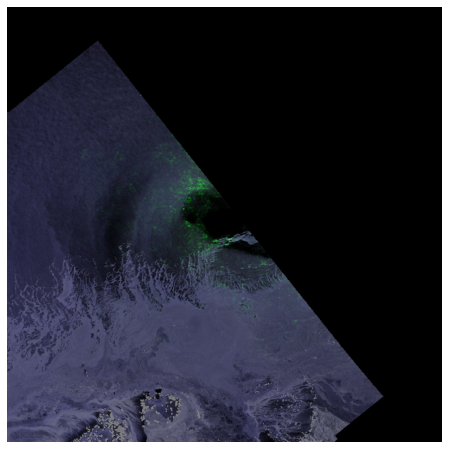}
        \includegraphics[width=0.2\textwidth]{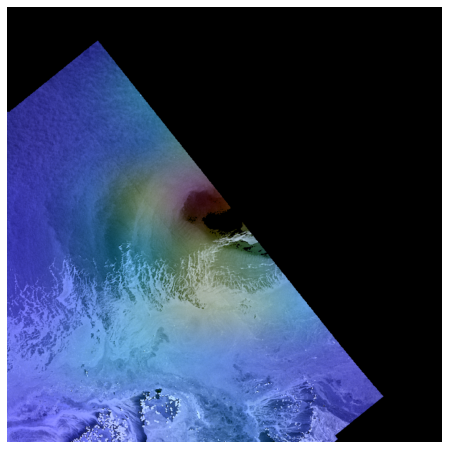}
    \\
        \includegraphics[width=0.2\textwidth]{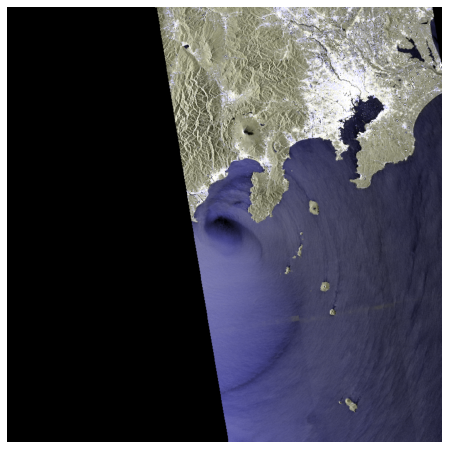}
        \includegraphics[width=0.2\textwidth]{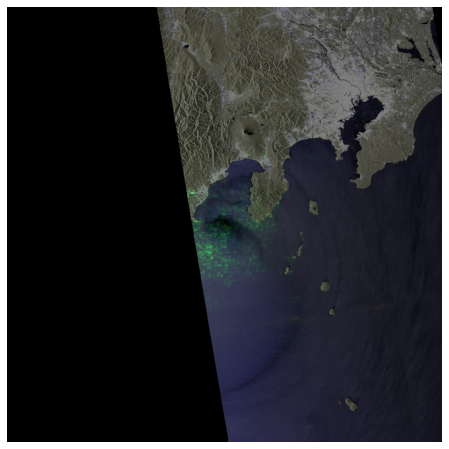}
        \includegraphics[width=0.2\textwidth]{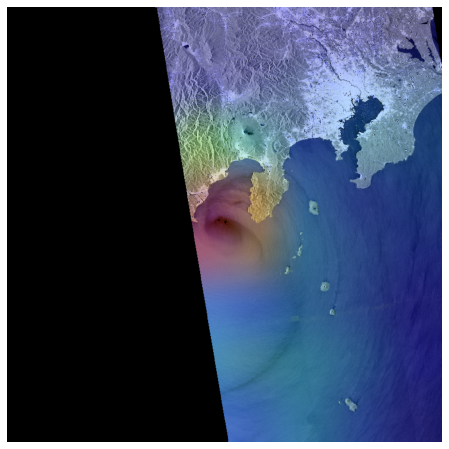}
    \\
        \includegraphics[width=0.2\textwidth]{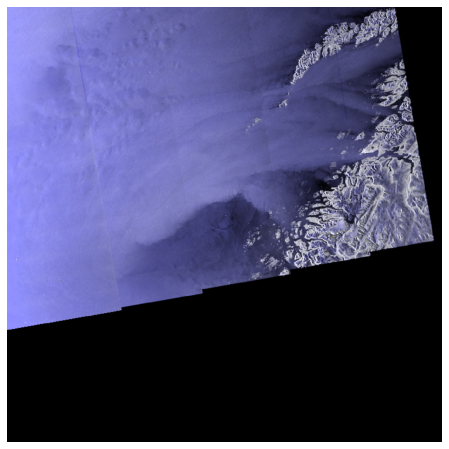}
        \includegraphics[width=0.2\textwidth]{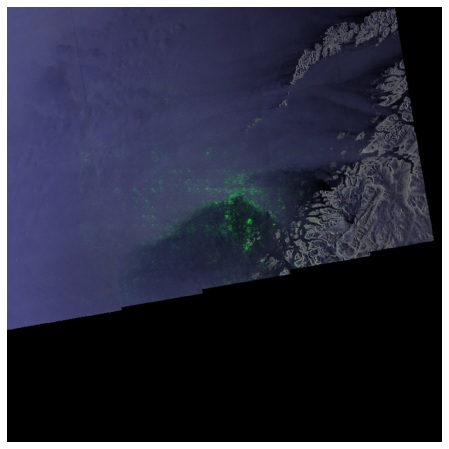}
        \includegraphics[width=0.2\textwidth]{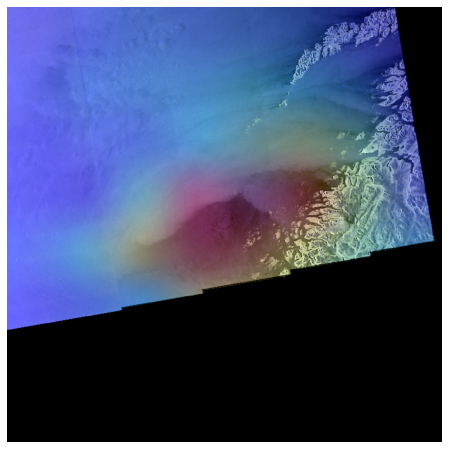}
    \\
        \includegraphics[width=0.2\textwidth]{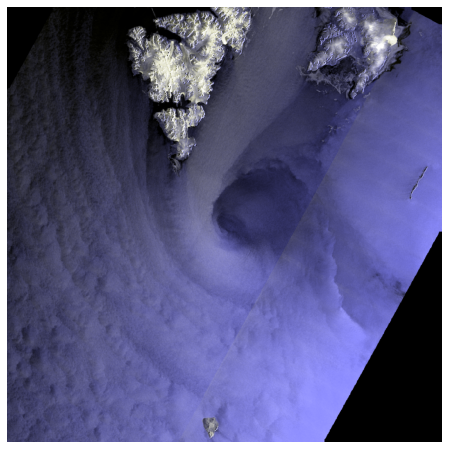}
        \includegraphics[width=0.2\textwidth]{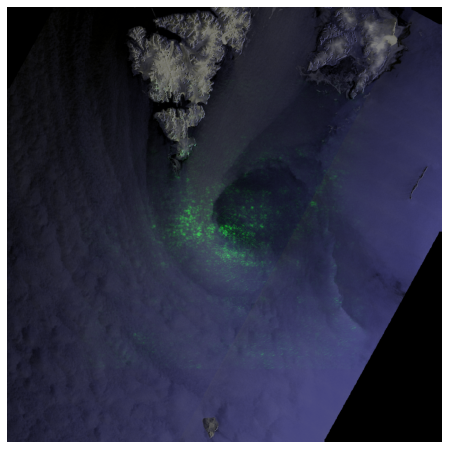}
        \includegraphics[width=0.2\textwidth]{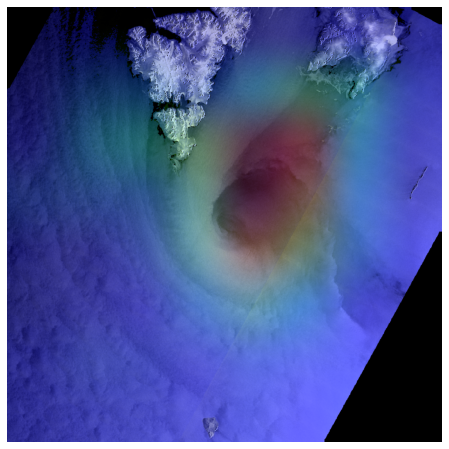}
    \caption{ True positives (4 of 62 samples shown) in first column and corresponding \gls{ig} and \gls{gcam} in the second and third column. The classification scores are 0.99, 0.82, 0.99, and 0.99, respectively. The polarised channels are HH/HV, HH/HV, VV/VH and HH/HV, respectively.}
	\label{fig:TP}
\end{figure*}

\subsection{True positives}\label{sec:tp}
Of the 62 \gls{tp} samples (i.e.\ low pressures correctly classified as low pressures), 4 samples are displayed in figure \ref{fig:TP}. The first column shows the input \gls{rgb} colour composites, the second column shows the \gls{ig} in green and the third column shows the \gls{gcam} as a heat map. 3 out of these 4 samples are located in polar regions, while the sample on the second row is an extra-tropical cyclone observed off the coast of Japan. The \gls{ig} and \gls{gcam} overlays indicate that the model is focusing on the cyclonic eye features. The \gls{ig} overlay has a slightly higher emphasis on the wind fronts as compared to the \gls{gcam}. Both the \gls{ig} and \gls{gcam} indicate that the model is effectively disregarding land features as well as the sea ice features appearing in the top row. Notably, in the top row, a large part of the cyclonic eye feature is also cropped due to the limited swath width of the \gls{sar}. This is the case in multiple samples classified as \gls{tp}, indicating a certain robustness to image features being cropped or obscured by e.g.\ sea ice.

\begin{figure*}[ht!]
\captionsetup[subfigure]{labelformat=empty}
	\centering
        \includegraphics[width=0.2\textwidth]{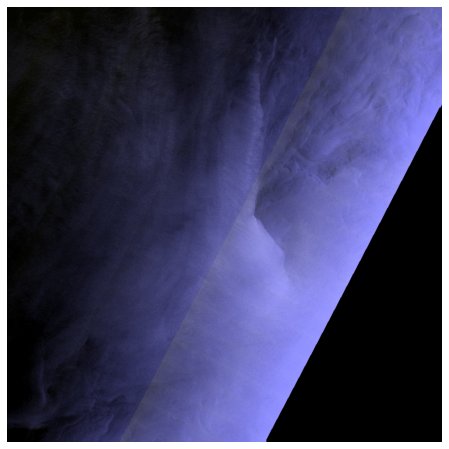}
        \includegraphics[width=0.2\textwidth]{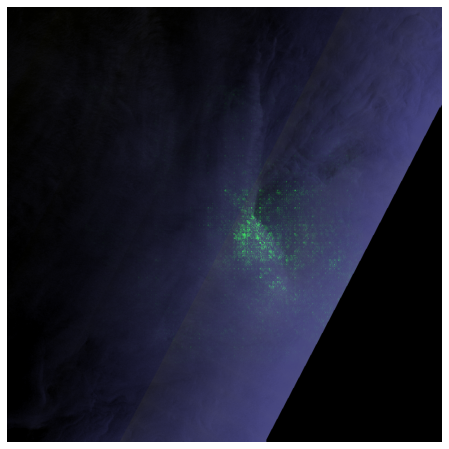}
        \includegraphics[width=0.2\textwidth]{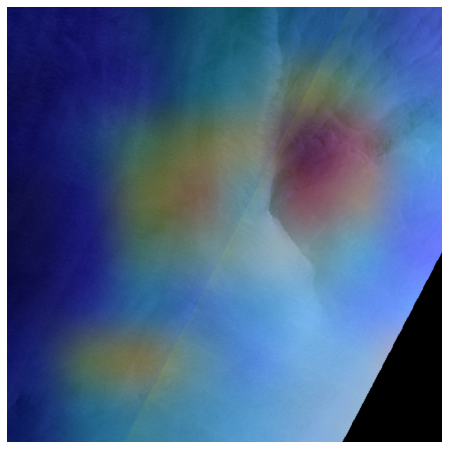}
    \\
        \includegraphics[width=0.2\textwidth]{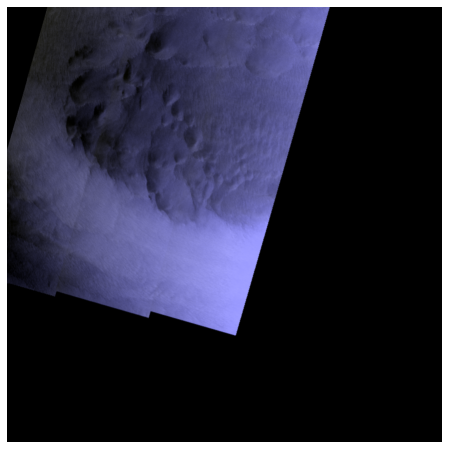}
        \includegraphics[width=0.2\textwidth]{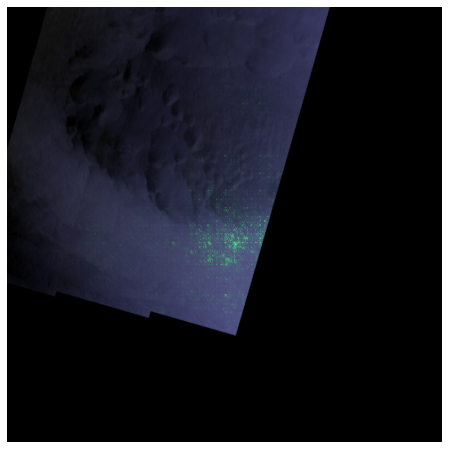}
        \includegraphics[width=0.2\textwidth]{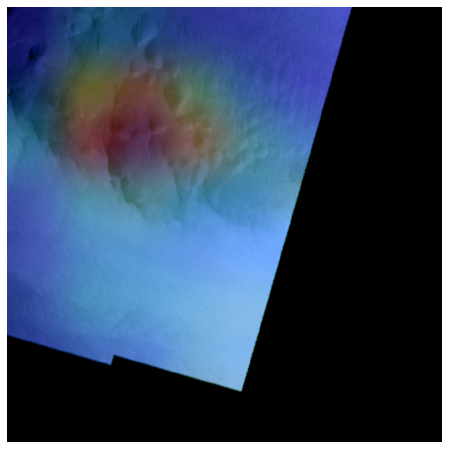}
    \\
        \includegraphics[width=0.2\textwidth]{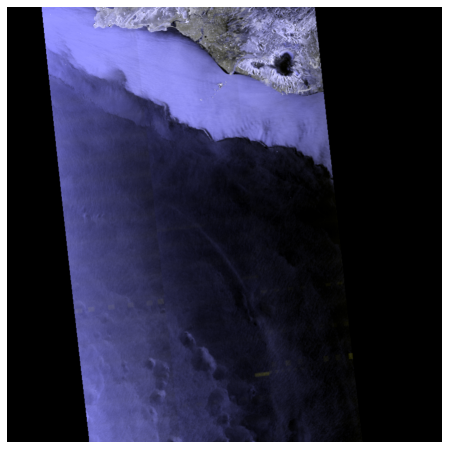}
        \includegraphics[width=0.2\textwidth]{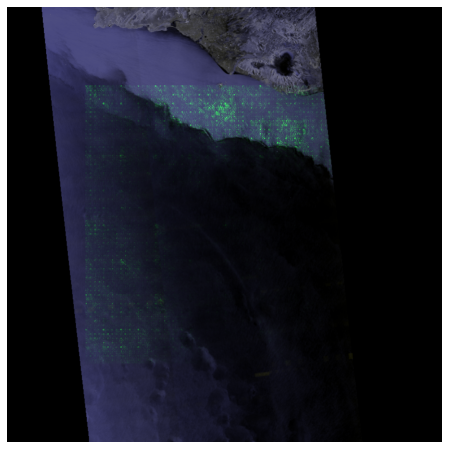}
        \includegraphics[width=0.2\textwidth]{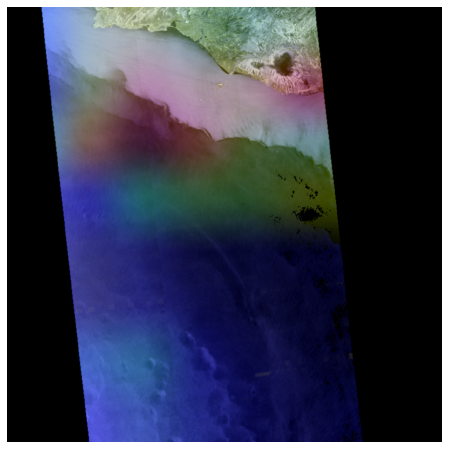}
    \\
        \includegraphics[width=0.2\textwidth]{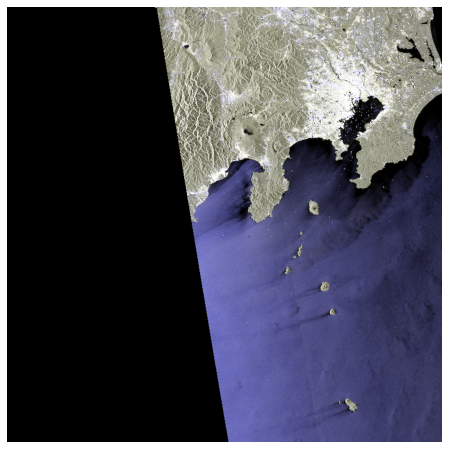}
        \includegraphics[width=0.2\textwidth]{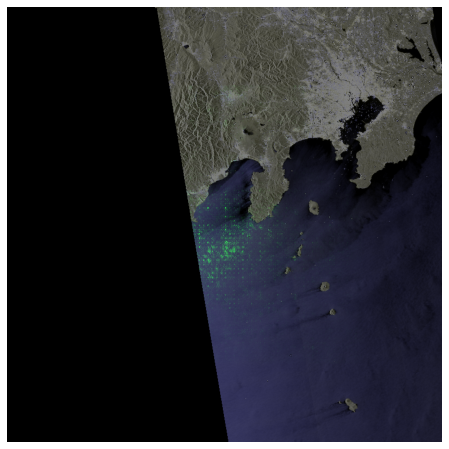}
        \includegraphics[width=0.2\textwidth]{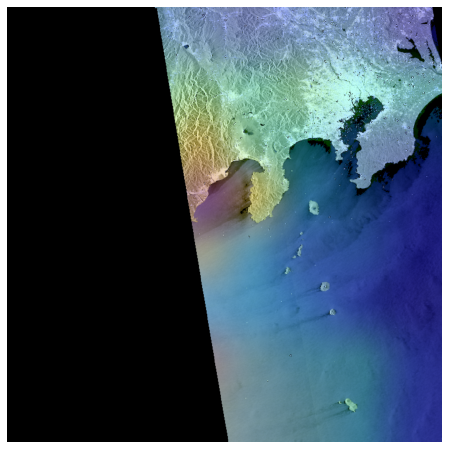}
    \caption{ False positives (4 of 5 samples shown), with classification scores 0.89, 0.92, 0.57 and 0.87. The polarised channels are HH/HV, HH/HV, VV/VH and VV/VH, respectively.}
	\label{fig:FP}
\end{figure*}

\subsection{False positives}\label{sec:fp}
The model classified 5 samples as \gls{fp} (i.e.\ absence of low pressures incorrectly classified as low pressures), of which 4 are shown in figure \ref{fig:FP}. The top two samples are presumably difficult to classify correctly (or the ground truth label could potentially be wrong), as they actually contain some pronounced wind fronts. Considering the \gls{ig} and \gls{gcam}, indeed the model is focusing on these wind features. The sample on the third row also contains a pronounced wind front that the model is focusing on, but the front is not curved. The classification score is however only 0.57 for this sample. In the forth sample, no wind front is visible, but the \gls{ig} and \gls{gcam} reveal that the model focuses on a wind wake (formed behind the Izu peninsula, Japan, located in the image centre), which may be misinterpreted as a cyclonic eye. 

Finally, we notice that \gls{ig} and \gls{gcam} highlight different areas in the second and third image.
Explainability techniques for deep learning are tools meant for diagnostic, which require a certain degree of subjective interpretation. Each technique is based on specific heuristics, which put a bias on what features are considered relevant. Indeed, even for samples classified with high confidence two explainability techniques might focus on different input features~\cite{samek2019explainable}. The discrepancy is often exacerbated in samples classified with lower confidence.

\begin{figure}[ht!]
\captionsetup[subfigure]{labelformat=empty}
	\centering
        \includegraphics[width=0.23\textwidth]{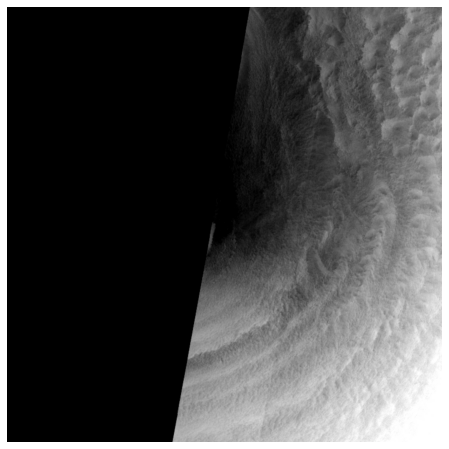}
        \includegraphics[width=0.23\textwidth]{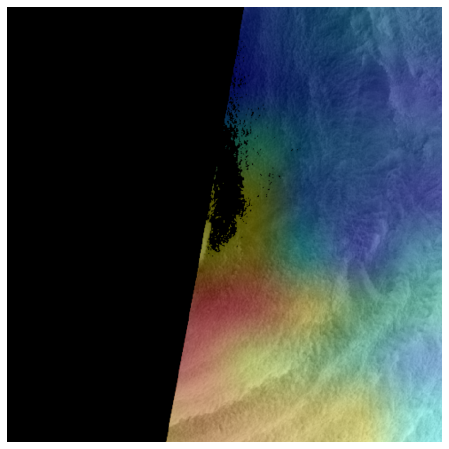}
    \\
        \includegraphics[width=0.23\textwidth]{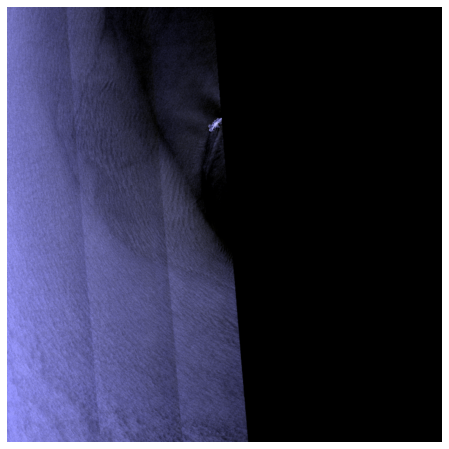}
        \includegraphics[width=0.23\textwidth]{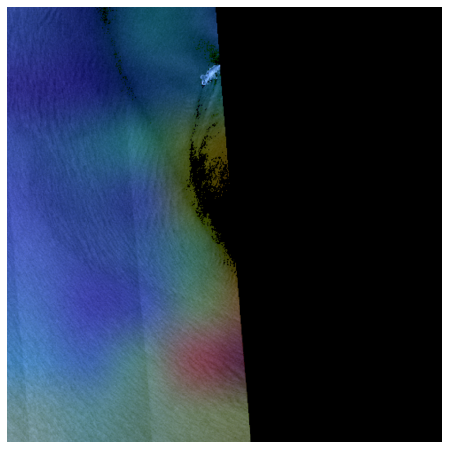}
    \caption{ False Negatives (all samples shown), with classification scores 0.96 and 0.79. The top image is acquired with the HH polarisation channel and the bottom image in the HH/HV polarised channels.}
	\label{fig:FN}
\end{figure}

\subsection{False negatives}\label{sec:fn}
Only two samples of the positive class were incorrectly classified as negatives, i.e.\ mistaken as absence of low pressure while being labeled as low pressures. These are shown in figure \ref{fig:FN}. Here, \gls{ig} are not computed, since a black image cannot be used as a baseline for the negative class. Nevertheless, \gls{gcam} can still be computed and is shown in the second column. It can be noted that both images suffer from lacking data due to the limited swath width of the \gls{sar} acquisitions. Indeed, the \gls{gcam} indicate that the model is not focusing on the darker center features as was the case for the \gls{tp} samples in figure \ref{fig:TP}. It should however be emphasised that this happens for only 2 of the 368 negative samples in the test set.

%% file: parts/conclusions.tex
\section{Conclusions}\label{sec:conclusions}
In this study, we show that \gls{sar} images from the Sentinel-1 satellites provide an attractive data source for automatic and accurate detection of maritime mesocyclones, such as polar lows. 
Specifically, we show that sufficiently many image examples can be found to build a labeled dataset for a deep learning model to be trained. By further comparing our deep learning model when trained on different input image resolutions, we find that higher resolution yields significantly better performance. This highlights the added value of using \gls{sar} data, as compared to conventionally used sensors of lower resolution. In particular, at 500 meters resolution we get an F1 score of 0.94, as compared to 0.87 at 2 km resolution (comparable to modern scatterometers). 

It should further be noted that the highest resolution tested in this study (500 meters) is primarily limited by the size of the training dataset and not the native resolution of the \gls{sar} sensor (10-40 meters). Thus, even higher input image resolutions could in principle be considered, potentially with even better performance. 
Larger input image sizes, however, ideally require deeper neural network architectures, with more trainable weights. This in turn require larger training datasets to avoid over-fitting. Even so, with an increasing amount of \gls{sar} data being available from new satellites, larger training datasets could be constructed in the future, enabling even better performance. 

By design, the training dataset contains spiral-form low pressures in the positive class. By analyzing \gls{ig} and \gls{gcam} on the trained model, we verify that the spiral shaped atmospheric fronts and the low wind centres yield most of the class attribution. 
Moreover, we conclude that:
\begin{enumerate*}[label=(\roman*)]
    \item these characteristic wind features do not need to be fully covered in the images, but can be substantially cropped due to the limited swath width of the \gls{sar}, 
    \item wind features can be partly covered by sea ice and still be identified by the model, and 
    \item the model is able to ignore land features in the images. 
\end{enumerate*}
The last point can be verified thanks to the procedure used to obtain the negative samples, i.e.\ through repeat-pass acquisitions (see section \ref{sec:negatives}). 


In summary, we conclude that the application of deep learning on \gls{sar} images for recognising matitime mesocyclones is promising. Further evaluation and comparison to detection based on data from other sensors or \gls{nwp} models is encouraged as a future work direction.

%% file: parts/acknowledgments.tex
\section{Acknowledgments}
This work was funded by the \gls{esa}, under the open call project \emph{Polar low detection based on Sentinel-1 data} (contract number 4000129961).
We would like to thank Patrick Stoll for his valuable feedback. We thank those involved in developing the GDAR software used to process \gls{sar} data, especially Heidi Hindberg, Yngvar Larsen, and Tom Grydeland. We also thank Temesgen Gebrie Yitayew and Hannah Vickers for their help in establishing this project. 
Finally, we would like to thank the reviewers for their insightful comments.

%% file: parts/supplementary.tex
\onecolumn
\setcounter{section}{0}

\begin{center}
    \huge{\textbf{Supplementary material}}
\end{center}

\section{Comparison with other deep learning architectures}

Here, we report the results achieved with off-the-shelf deep learning architectures for image classification.
In Tab.~\ref{tab:other_arch} we report the results obtained with VGG16~\cite{DBLP:journals/corr/SimonyanZ14a}, ResNet50~\cite{he2016deep}, Xception~\cite{chollet2017xception}, MobileNet~\cite{DBLP:journals/corr/HowardZCKWWAA17}, ViT~\cite{DBLP:conf/iclr/DosovitskiyB0WZ21}, and MLPMixer~\cite{tolstikhin2021mlp}.
The implementation of VGG16, ResNet50, Xception, and MobileNet is the one from \textit{Keras applications}\footnote{\url{https://keras.io/api/applications/}}.

\begin{table}[!ht]
\centering
\footnotesize	
\begin{tabular}{c|ccccc}
Architecture & TN & FN & FP & TP & F1 score \\ \hline
VGG16     & 371 & 64 & 0  & 0  & 0.0  \\ %
ResNet50  & 366 & 5  & 11 & 53 & 0.87 \\ %
Xception  & 365 & 4 & 6  & 60 & 0.92 \\ %
MobileNet & 369 & 11 & 2  & 53 & 0.89 \\ %
ViT       & 350 & 9  & 21 & 55 & 0.79 \\ %
MLPMixer  & 306 & 25 & 65 & 39 & 0.46 \\ %
\hline
\end{tabular}
\caption{ Results obtained with popular architectures. The best performance in terms of F1 score obtained across 5 independent runs is reported.}
\label{tab:other_arch}
\end{table}

From the Table, we see that the best performance are obtained by Xception and MobileNet, the two popular deep learning architectures using separable 2D convolutions. Such a result, encouraged us to adopt in our experiments an architecture with residual connections and SepConv2D layers, similar to Xception and MobileNet.

\section{Comparison of different techniques to handle class imbalance}

Tab.~\ref{tab:class_imbalance} reports the classification performance and training times when using different techniques to handle class imbalance. Despite being more computationally intensive, the oversampling technique yields the best classification performance and, thus, is the one adopted in the experimental evaluation.

\begin{table}[!ht]
\centering
\footnotesize	
\begin{tabular}{c|cccccc}
Balancing method   & Time/epoch & TN & FN & FP & TP & F1 score \\ \hline
Class-weighting    & 18s & 369.4{\tiny$\pm$2.1} & 12.3{\tiny$\pm$1.6} & 2.3{\tiny$\pm$1.3} & 52.1{\tiny$\pm$1.9} & 0.88{\tiny$\pm$0.01} \\ %
Oversampling       & 31s & 367.8{\tiny$\pm$2.7} & 3.2{\tiny$\pm$2.7} & 4.6{\tiny$\pm$1.7} & 59.4{\tiny$\pm$1.7} & 0.94{\tiny$\pm$0.01} \\ %
Rejection sampling & 22s & 367.3{\tiny$\pm$2.4} & 6.1{\tiny$\pm$1.4} & 4.2{\tiny$\pm$1.1} & 58.0{\tiny$\pm$2.1} & 0.92{\tiny$\pm$0.02} \\ %
\hline
\end{tabular}
\caption{ Classification performance obtained with different methods to handle class imbalance. }
\label{tab:class_imbalance}
\end{table}

\section{Optimal hyperparameters for models trained on lower resolution images}

We optimized the hyperparameters for the models trained on lower resolution by following the exact same procedure that we used for the model operating on the higher resolution images.
The optimal hyperparameters for the different models are reported in the Tab.~\ref{tab:hyperparams_lowres}.
We note that the optimal hyperparameters are the same for different image resolutions, except for: \textit{Num. residual blocks}, \textit{Dropout rate}, and \textit{Learning rate}.  

\begin{table}[!ht]
\footnotesize
\centering
\begin{tabular}{llll}
\hline
\textbf{Hyperparam.}              & \textbf{500m} & \textbf{1km} & \textbf{2km} \\ \hline
Activation                        & ReLU  & ReLU & ReLU  \\ 
Conv2D filters                    & 8 & 8 & 8      \\ 
Kernel size                       & 3 & 3 & 3       \\ 
SepConv2D filters                 & 8 & 8 & 8       \\ 
Num. residual blocks              & 7 & 5 & 4       \\ 
Global pooling                    & avg & avg & avg     \\ 
Num. dense layers                 & 1 & 1 & 1       \\ 
Units in the dense layer          & 8 & 8 & 8       \\ 
Dropout rate                      & 0.5 & 0.4 & 0.6     \\ 
Use batch normalization           & True & True & True    \\
Learning rate                     & 1e-3 & 1e-2 & 1e-3    \\ 
\hline
\end{tabular}
\caption{ Optimal hyperparameters for the models trained on different image resolutions.}
\label{tab:hyperparams_lowres}
\end{table}


\section{Comparison between co- and cross-polarised channels}
Here, results using only one polarisation channel are presented for the 500 m resolution images. Specifically, the F1-score is presented in table \ref{tab:single_channel} (averaged over 5 independent runs), obtained when using the co- or cross-polarised channels separately. For the co-polarised case, both HH and VV are used jointly, while for the cross-polarised chase, the VH and HV channels are used jointly. The results show that the cross-polarised channels yields better performance compared to the co-polarised channels. A possible reason for this could be that the cross-polarised channels better captures high wind speed features. At high wind speeds, the co-polarised backscatter saturates faster as a function of wind speed \cite{zhang2017hurricane}. Another factor that could play a role is that the two co-polarisations (HH and VV) typically behaves differently as a function of incidence angle and target properties, making the dataset somewhat heterogeneous. The cross-polarisation (VH and HV) dataset is on the other hand homogeneous, since, theoretically these channels are identical for reciprocal targets, like the ocean. 

The best performance is however still obtained using both polarisation channels, as shown in table \ref{tab:res_all_folds} in the main manuscript. 

\begin{table}[!ht]
\centering
\footnotesize	
\begin{tabular}{l|cc}
  & Co-pol ($x_{\rvert\rvert}$) &  Cross-pol ($x_{\times}$) \\ \hline
  F1-score & 0.886$\pm$0.016 & 0.916$\pm$0.013 \\
\hline
\end{tabular}
\caption{ Results obtained using only co- or cross-polarised channels separately.}
\label{tab:single_channel}
\end{table}

\section{Training time for different image resolutions}

Tab.~\ref{tab:train_times} reports the training times of the proposed deep learning architecture when images of different resolutions are used in training. The training times are measured on an Nvidia RTX 3090.
Clearly, lower resolutions result in a much faster training. However, even when using 500m resolution, the neural network can be trained reasonably fast.

About the differences in time for the inference phase, they are negligible when using different image resolutions (a fraction of a second in each case).
Considering the whole process from satellite acquisition, data downlink/download, SAR focusing, pre-processing (in particular geocoding) etc, the inference time of the neural network model is by all means negligible in an operational setting.

\begin{table}[!ht]
\centering
\footnotesize	
\begin{tabular}{c|ccc}
Pixel size  & 500m & 1km & 2km \\ \hline
Time/epoch  & 31s  & 9s  & 5s  \\ %
\hline
\end{tabular}
\caption{ Training times for different image resolutions.}
\label{tab:train_times}
\end{table}

\section{Interpretability for a model trained on low-resolution images}
An interesting question when comparing model performance between input image resolutions (500, 1000 and 2000 metres), is if the interpretability metrics (\gls{ig} and \gls{gcam}) can indicate why the performance is worse for the lower resolutions. By comparing the results of the high-res model trained on the 500m resolution images to the results of the low-res model trained on the 2000m resolution images, we find that the low-res model miss-classifies 11 \glspl{fn} and 7 \glspl{fp}. Among these, there are 9 \glspl{fn} and 5 \glspl{fp} that the high-res model classifies correctly. Therefore, we compare the low-res \glspl{fn} to corresponding high-res \glspl{tp}, and low-res \glspl{fp} to high-res \glspl{tn}. 

Since we cannot compute \gls{ig} on samples classified as negatives (as explained in section \ref{sec:fn}), we only consider the \gls{gcam}. It should however be noted that the \gls{gcam} heat map is a result of gradients at the last layer in the model (see section \ref{sec:gcam} for details). Since each layer in the model contains pooling, the heat map will be of lower resolution than the input image itself. The \gls{gcam} heat map thus provide little or no information about fine detailed differences. Yet, it is expected that differences between the high- and low-res results are primarily fine details (which disappear when the resolution is lowered). Despite this limitation in the analysis, differences in the interpretability results with regard to \gls{gcam} are presented below. 

\subsection{Low-res \glspl{fn} versus high-res \glspl{tp}}
\begin{figure}[ht!]
\captionsetup[subfigure]{labelformat=empty}
	\centering
        \includegraphics[width=0.40\textwidth]{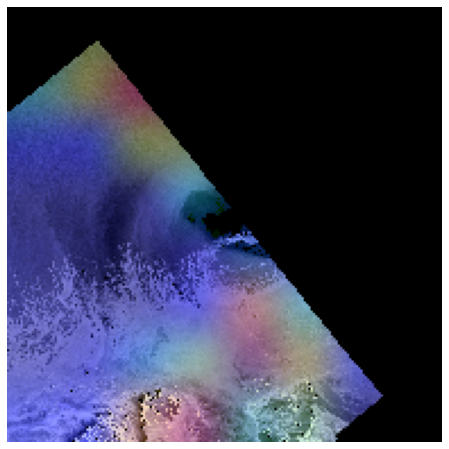}
        \includegraphics[width=0.40\textwidth]{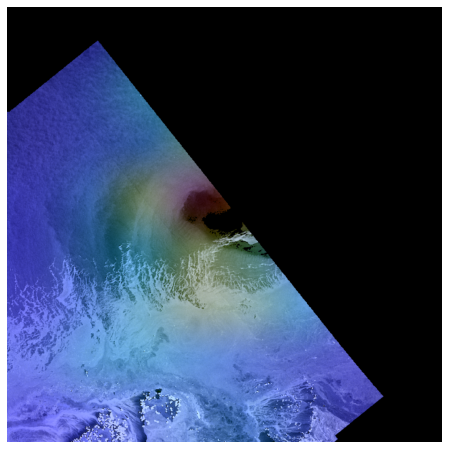}
    \\
        \includegraphics[width=0.40\textwidth]{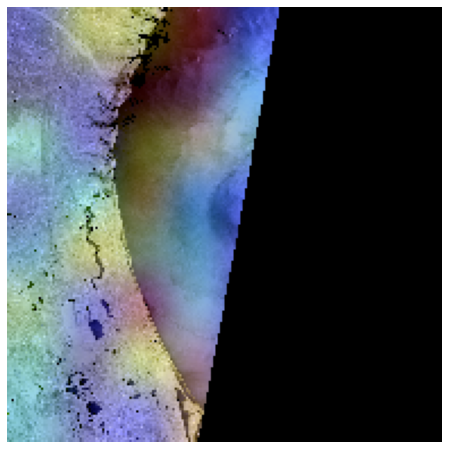}
        \includegraphics[width=0.40\textwidth]{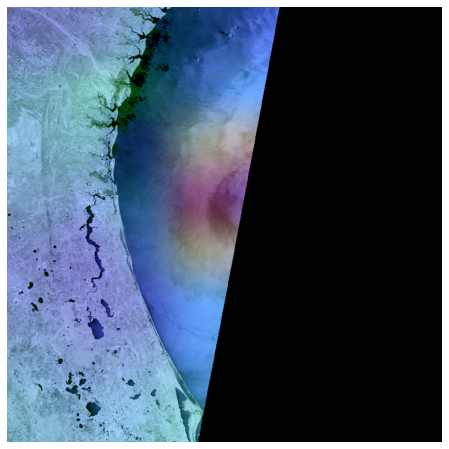}
    \\
        \includegraphics[width=0.40\textwidth]{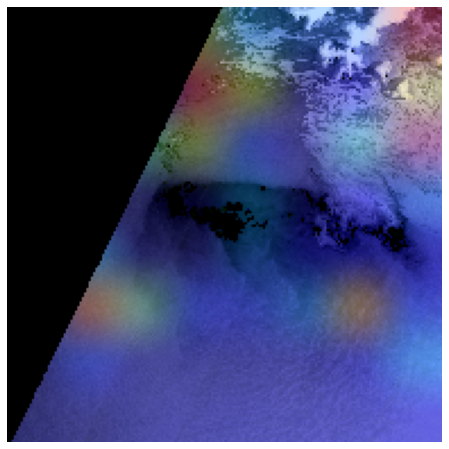}
        \includegraphics[width=0.40\textwidth]{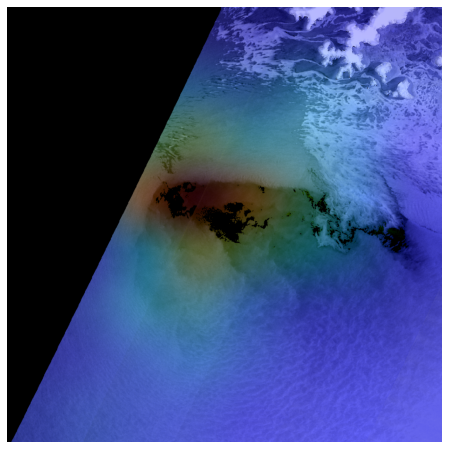}
    \caption{ \Gls{gcam} heat maps for the \glspl{fn} of the low-res model to the left, and the corresponding \gls{tp} of the high-res model to the right. }
	\label{fig:FN-TP}
\end{figure}

The \glspl{fn} of the low-res model (that the high-res model classifies correctly), could give insights into what key features of the input images are lacking at the lower resolution in order to correctly classify an image with a mesocyclone. The low-res \glspl{fn} and the corresponding high-res \glspl{tp} are shown in figure \ref{fig:FN-TP} (3 of the total of 9 cases are shown). It is clear that the low-res model does not attribute the same importance to the cyclonic eye or wind front features as the high-res model. This could indicate that at the cyclonic eye or at the wind front, high-res features are of particular importance. If these are lacking, the model focuses elsewhere in the image. In the shown examples, the attribution of the low-res model appears considerably more scattered, which could indicate that there are not sufficiently strong features to attract the model attention. 

\subsection{Low-res \glspl{fp} versus high-res \glspl{tn}}
\begin{figure}[ht!]
\captionsetup[subfigure]{labelformat=empty}
	\centering
        \includegraphics[width=0.40\textwidth]{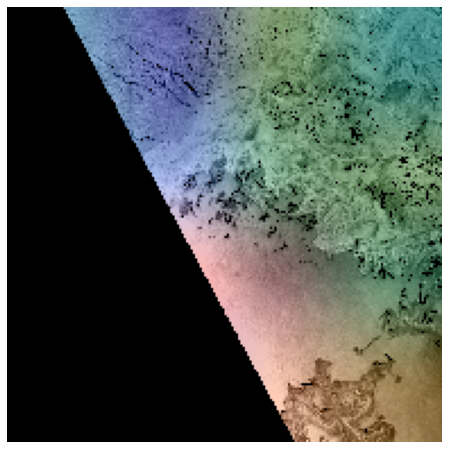}
        \includegraphics[width=0.40\textwidth]{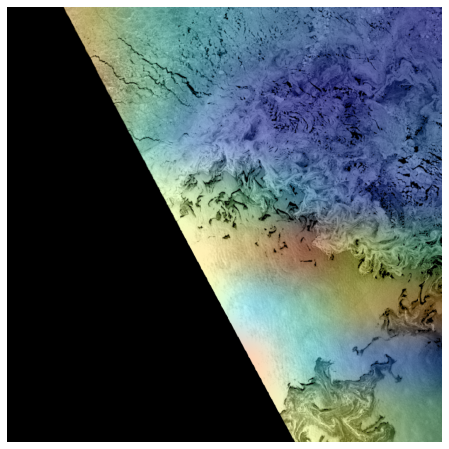}
        \includegraphics[width=0.40\textwidth]{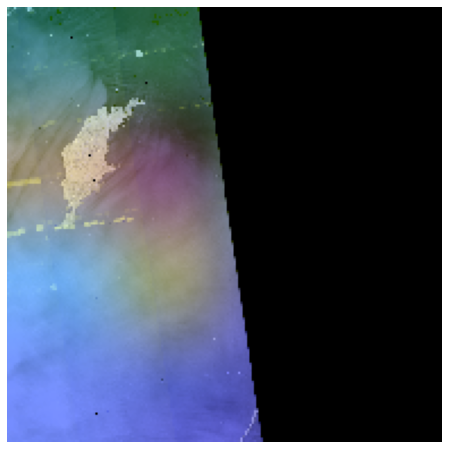}
        \includegraphics[width=0.40\textwidth]{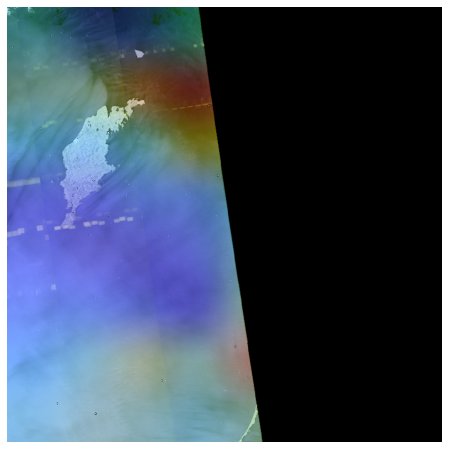}
    \\
        \includegraphics[width=0.40\textwidth]{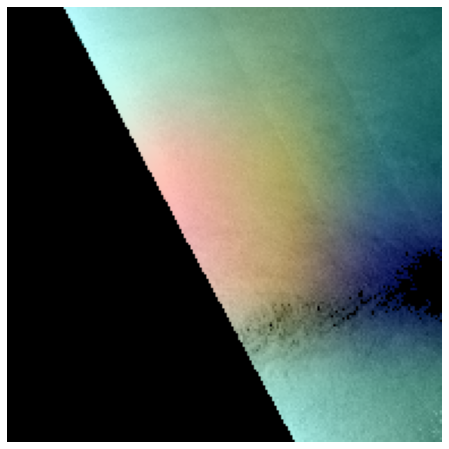}
        \includegraphics[width=0.40\textwidth]{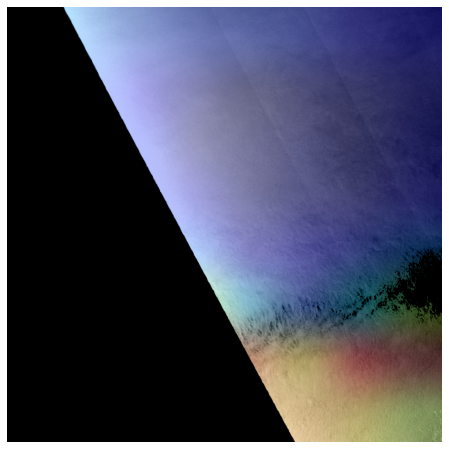}
    \\
    \caption{ \Gls{gcam} heat maps for the \glspl{fp} of the low-res model to the left, and the corresponding \gls{tn} of the high-res model to the right. }
	\label{fig:FP-TN}
\end{figure}

The low-res \glspl{fp} and the corresponding high-res \glspl{tn} are shown in figure \ref{fig:FP-TN} (3 of the total of 5 cases are shown). It is clear that the low-res model now puts the main attribution to the image centres, while the attribution of the high-res model is more scattered. In the top sample, part of the image is covered by sea ice, which is rich of fine details. At the lower resolution, these features could potentially be sufficiently blurred for the low-res model to be confused, e.g., with a wind front. In the second sample, a slight wind front seems to be picked up, but it is unclear what the low-res model actually is focusing on. The situation is similar in the third example, where no clear feature is shown at the centre.

